\documentclass[10pt,twocolumn,letterpaper]{article}

\usepackage[pagenumbers]{cvpr} 

\usepackage{graphicx}
\usepackage{amsmath}     
\usepackage{multirow}
\usepackage{amssymb}
\usepackage{colortbl}
\usepackage{wrapfig}
\usepackage[most]{tcolorbox}
\usepackage{makecell}
\usepackage{xspace}
\usepackage{subcaption}
\usepackage{enumitem}
\usepackage{pifont}
\usepackage{color, colortbl}

\newcommand{\ours}{FusionAgent\xspace}
\newcommand{\fusion}{ACT\xspace}
\newcommand{\da}{Direct Answering\xspace}
\newcommand{\think}{\textit{$<$think$>$...$<$/think$>$}\xspace}

\newcommand{\answer}{\textit{$<$answer$>$...$<$/answer$>$}\xspace}

\renewcommand{\paragraph}[1]{\vspace{1.25mm}\noindent\textbf{#1}}

\def\eg{\emph{e.g}\onedot} 
\def\ie{\emph{i.e}\onedot} 
 
 \def\vs{\emph{vs}\onedot}
\def\wrt{w.r.t\onedot}

\makeatother

\definecolor{green}{HTML}{0aa344}
\definecolor{red}{HTML}{c93756}
\definecolor{fullgreen}{rgb}{0.502, 0.788, 0.643}
\definecolor{fullred}{rgb}{0.800, 0.447, 0.541}
\definecolor{lightgreen}{RGB}{225, 239, 217}   
\definecolor{lightblue}{RGB}{203, 220, 235}   
\definecolor{fullgray}{RGB}{219, 223, 234}   
\definecolor{fullpurple}{RGB}{205, 193, 255}
\definecolor{darkred}{RGB}{204, 114, 138}
\definecolor{darkpurple}{RGB}{171, 151, 255}
\definecolor{darkgray}{RGB}{114, 114, 114}
\definecolor{teal}{HTML}{14B8A6}
\definecolor{sky}{HTML}{38BDF8}
\definecolor{indigo}{HTML}{6366F1}
\definecolor{navy}{HTML}{1E3A8A}
\definecolor{amber}{HTML}{F59E0B}
\definecolor{coral}{HTML}{FF6B6B}
\definecolor{peach}{HTML}{FFB4A2}
\definecolor{sage}{HTML}{A8C3A1}
\definecolor{dustyblue}{HTML}{9BB4C7}
\definecolor{mauve}{HTML}{BFA6C9}
\definecolor{clay}{HTML}{C9B29B}
\newcommand{\cmark}{\textcolor{fullgreen}{\textbf{\checkmark}}}
\newcommand{\xmark}{\textcolor{fullred}{\textbf{\ding{55}}}}

\definecolor{cvprblue}{rgb}{0.21,0.49,0.74}
\usepackage[pagebackref,breaklinks,colorlinks,allcolors=cvprblue]{hyperref}

\def\paperID{} 
\def\confName{CVPR}
\def\confYear{2026}

\title{\ours: A Multimodal Agent with Dynamic Model Selection for \\ Human Recognition}

\author{
Jie Zhu\textsuperscript{1}\quad Xiao Guo\textsuperscript{1}\quad Yiyang Su\textsuperscript{1}\quad Anil Jain\textsuperscript{1}\quad Xiaoming Liu\textsuperscript{1,2} \\
\textsuperscript{1}Michigan State University \quad \textsuperscript{2}University of North Carolina at Chapel Hill\\
{\tt\small \{zhujie4, guoxia11, suyiyan1, jain\}@msu.edu \quad liuxm@cs.unc.edu }
}

\begin{document}
\maketitle


\begin{abstract}

Model fusion is a key strategy for robust recognition in unconstrained scenarios, as different models provide complementary strengths. This is especially important for whole-body human recognition, where biometric cues such as face, gait, and body shape vary across samples and are typically integrated via score-fusion. However, existing score-fusion strategies are usually static, invoking all models for every test sample regardless of sample quality or modality reliability.
To overcome these limitations, we propose \textbf{\ours}, a novel agentic framework that leverages a Multimodal Large Language Model (MLLM) to perform dynamic, sample-specific model selection. Each expert model is treated as a tool, and through Reinforcement Fine-Tuning (RFT) with a metric-based reward, the agent learns to adaptively determine the optimal model combination for each test input. To address the model score misalignment and embedding heterogeneity, we introduce Anchor-based Confidence Top-k (ACT) score-fusion, which anchors on the most confident model and integrates complementary predictions in a confidence-aware manner. Extensive experiments on multiple whole-body biometric benchmarks demonstrate that \ours significantly outperforms SoTA methods while achieving higher efficiency through fewer model invocations, underscoring the critical role of dynamic, explainable, and robust model fusion in real-world recognition systems. Project page: \href{https://fusionagent.github.io/}{FusionAgent}.

\end{abstract}

\begin{figure}[t!]
    \centering
    \includegraphics[width=\linewidth]{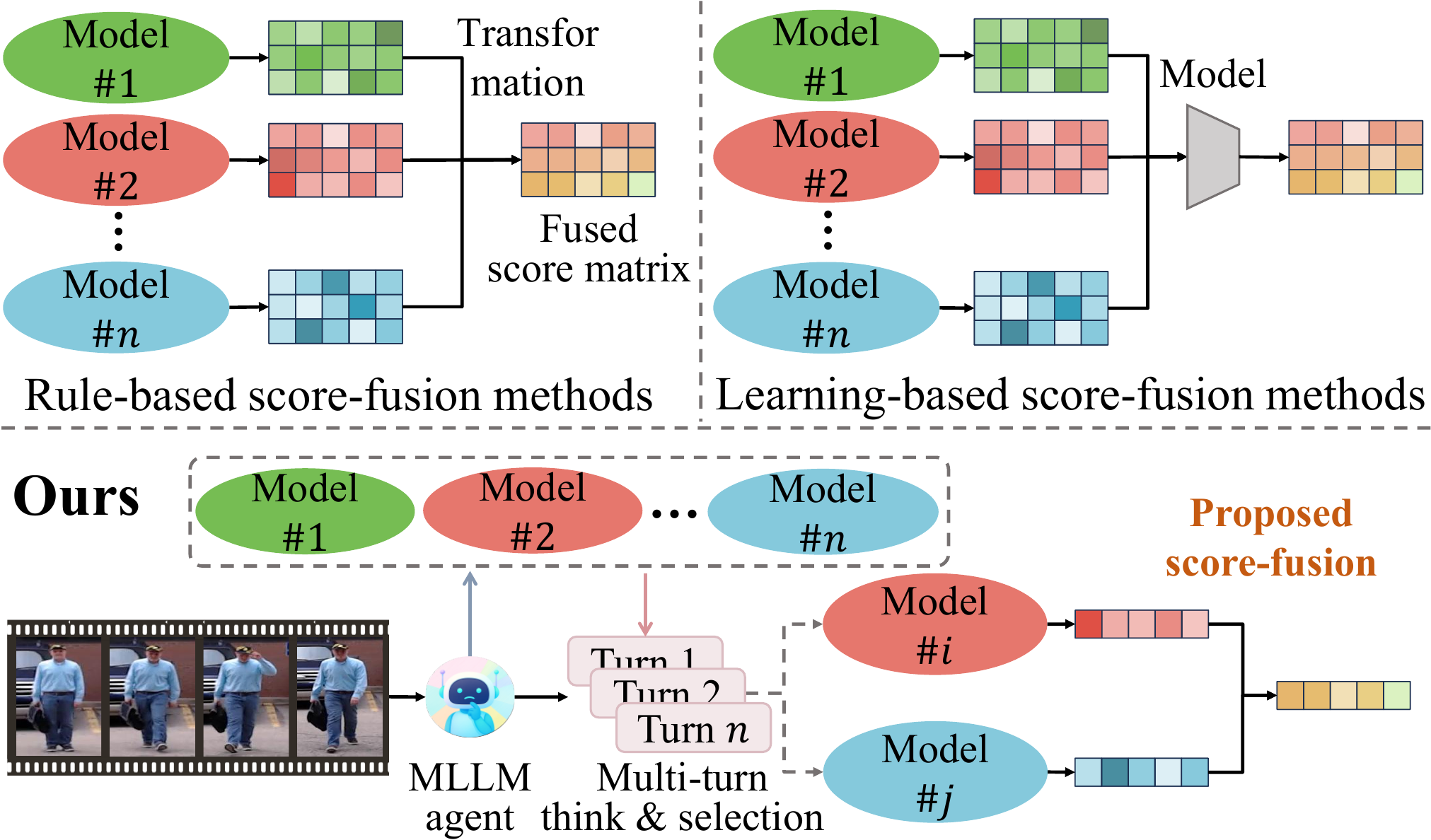}
    \caption{\textbf{Comparison of score-fusion methods. Top:} Rule-based methods apply predefined transformations to fuse all model scores, while learning-based methods infer a fusion model from data but still assume that every model contributes to all test samples. \textbf{Bottom:} our framework leverages an MLLM agent to dynamically select a subset of models, followed by the proposed score-fusion strategy, enabling adaptive and robust integration.}
    \label{fig:motivation}
\vspace{-0.5em}
\end{figure}

\section{Introduction} \label{sec:intro}

Computer vision has made remarkable progress through state-of-the-art models for tasks such as object detection, segmentation, face recognition, and image classification~\cite{chen2023atm, kim2025sapiensid, kumar2025charm3r, guo2025rethinking, zhang2026chain, huang2026unlocking, zhang2026towards, guo2026holistic}. Yet existing research has largely focused on single-model performance, paying far less attention to how complementary models can be effectively combined. Since different models often excel under different conditions, selective fusion offers a promising path toward stronger and more robust performance.

Model fusion is widely used with applications ranging from feature fusion to score fusion~\cite{zhang2023tile, snelick2003multimodal, zhang2024tamm}. Among these, score fusion is particularly representative and practical, especially in biometric recognition~\cite{jain2005score, park2021multi, zhu2025quality}. As shown in Fig.~\ref{fig:motivation}, existing methods are mainly rule-based or learning-based. The former combine scores with predefined rules~\cite{snelick2003multimodal, jain2005score}, while the latter learn fusion strategies from data~\cite{park2021multi, zhu2025quality, teng2022optimized}. Despite their differences, both typically rely on a fixed combination of models for all inputs, assuming universal complementarity. This assumption is often violated in practice: for example, face recognition scores are unreliable when only a person’s back view is visible. Even methods such as QME~\cite{zhu2025quality}, which account for input quality, still allow low-content inputs to affect the final output. These limitations motivate a more selective fusion paradigm that first determines which models are suitable for each input, and then how to combine their outputs effectively. To systematically study this problem, we pose the central research question of this paper:

\begin{tcolorbox}[colback=lightgray!55, colframe=lightgray!75!black, boxrule=0pt, arc=2mm, left=1mm, right=1mm, top=1mm, bottom=1mm]
\textit{How can we adaptively choose the \textbf{optimal model combination for each test sample} while rigorously testing whether such selection improves performance even with an \textbf{ad hoc score-fusion algorithm}?}
\end{tcolorbox}

This raises the question of how to perform sample-wise model selection in a principled and interpretable manner. Recent advances in multimodal large language models (MLLMs) suggest a promising direction for adaptive model fusion~\cite{yao2023react, liu2025arft, chen2025lvagent, zuo20254kagent, xiao2026deepfake}. By reasoning over complex inputs and orchestrating external tools in a context-dependent manner, an agent can treat each specialized model as a tool and decide which ones to invoke for each sample. This makes agentic fusion particularly suitable for selective fusion, where model reliability varies across inputs and decisions must be made adaptively for each sample.

This motivation is especially compelling in systematic whole-body recognition~\cite{zhu2025quality}, which integrates multimodal traits (\eg, face, gait, and body shape) and multiple specialized models. Unlike unimodal systems based on face recognition (FR)~\cite{deng2019arcface, kim2022adaface, kim2024keypoint}, gait recognition (GR)~\cite{zhang2019gait, ye2024biggait}, or person re-identification (ReID)~\cite{gu2022clothes, liu2024distilling, su2025hierarchical}, whole-body recognition depends on complementary modalities whose reliability varies with input conditions. For example, face cues may fail under occlusion or extreme viewpoints, while gait or body appearance remains informative~\cite{liu2024farsight}. This makes whole-body recognition a representative testbed for agentic model selection and fusion.

Inspired by these, we introduce a unified framework centered around two core components, each designed to systematically respond to model selections and score-fusion methods. In response to model selections, we propose \textbf{\ours}, an MLLM-based agent that performs dynamic model selection per sample. Each biometric model is wrapped as a tool, providing the agent with a score vector and predicted label. Through Reinforcement Fine-Tuning (RFT) guided by the \textbf{proposed metric-based reward}, the agent learns to identify sample-specific model combinations by analyzing query patterns, effectively answering how optimal ensembles can be found for each input. To explore score-fusion methods, we design a lightweight \textbf{Anchor-based Confidence Top-k score-fusion method (\fusion)}. This method leverages the most confident model selected by the agent, then integrates only the top‑matching scores scaled by confidence weights. By doing so, \fusion mitigates misalignment from sample-wise model selection and embedding heterogeneity, demonstrating that even a simple fusion rule can yield strong performance when guided by adaptive model selection. Our main contributions are:

\begin{itemize}

\item We propose \ours, an agentic framework that leverages an MLLM to perform explainable, sample-wise dynamic model selection.

\item We introduce an Anchor-based Confidence Top-k score-fusion method (\fusion) to mitigate score misalignment arising from dynamic model selection, ensuring robust integration of heterogeneous score outputs.

\item We design a metric-based reward function, central to our optimization, which directly aligns the agent's selection strategy with final performance metrics.

\item Extensive experiments on multiple whole-body biometric benchmarks demonstrate the superiority of our approach over state-of-the-art methods, even using conventional score-fusion methods.

\end{itemize}

\section{Related Work} \label{sec:related work}

\subsection{Whole-body Biometric Recognition}

Whole-body biometric recognition systems combine detectors (\eg, for whole body and face), embedding models, and fusion modules to leverage multi-modal cues such as face and body features~\cite{de2012cabala, liu2024farsight, liu2025person,su2026localscore}. The key challenge is to effectively integrate the complementary strengths of different modalities and their dedicated models to maximize overall performance. For example, FR models excel with high-quality frontal faces but struggle under adverse conditions like oblique angles~\cite{deng2020sub, kim2022cluster, kim2022adaface, kim2024keypoint}. In contrast, GR models focus on clothing-invariant dynamic body attributes~\cite{zhang2020learning, ye2024biggait, zhang2019gait, ye2025biggergait, huang2026unlocking}, while ReID models take a holistic approach to extract comprehensive appearance features~\cite{wu2020adaptive, gu2022clothes, liu2023learning, liu2024distilling, yang2023good}. Prior fusion methods typically use all available models for every query~\cite{liu2024farsight, liu2025person, zhu2025quality, teng2022optimized, cheniti2024approach, park2021multi, ji2026idselect}, ignoring the sample-dependent nature of optimal model combinations. A clear case is a low-resolution image with a side-view of a person: FR models may be unreliable, making GR or ReID models more critical for that specific sample. 
Therefore, we propose \ours to dynamically select an optimal model combination for each test sample through explainable analysis, thereby tailoring the fusion process to individual inputs.

\subsection{Score-fusion}

Unlike feature-level fusion~\cite{kim2025sapiensid,arevalo2017gated, perez2019mfas, poh2011unified, su2025hierarchical, zhang2023tile}, score-level fusion integrates similarity scores from multiple modalities to improve recognition performance~\cite{singh2019comprehensive}. This approach is broadly categorized into two paradigms: rule-based and learning-based methods. Rule-based methods employ fixed rules such as Z-score, Min-max normalization, max/min fusion~\cite{ross2003information,jain2005score,yilmaz2016score, jain2011introduction}, and likelihood ratio-based fusion~\cite{poh2007improving,nandakumar2007likelihood,vatsa2007integrating,poh2011unified,he2010performance}, offering simplicity and efficiency. In contrast, learning-based methods optimize fusion during training~\cite{teng2022optimized,cheniti2024approach,park2021multi}. These include the recent quality-aware approach QME~\cite{zhu2025quality}, which performs weighted score-fusion based on input quality. In contrast, we introduce a simple score-fusion algorithm with an anchor. This design, combined with selective model inputs, significantly outperforms existing fusion strategies.

\begin{figure*}
    \centering
    \includegraphics[width=\linewidth]{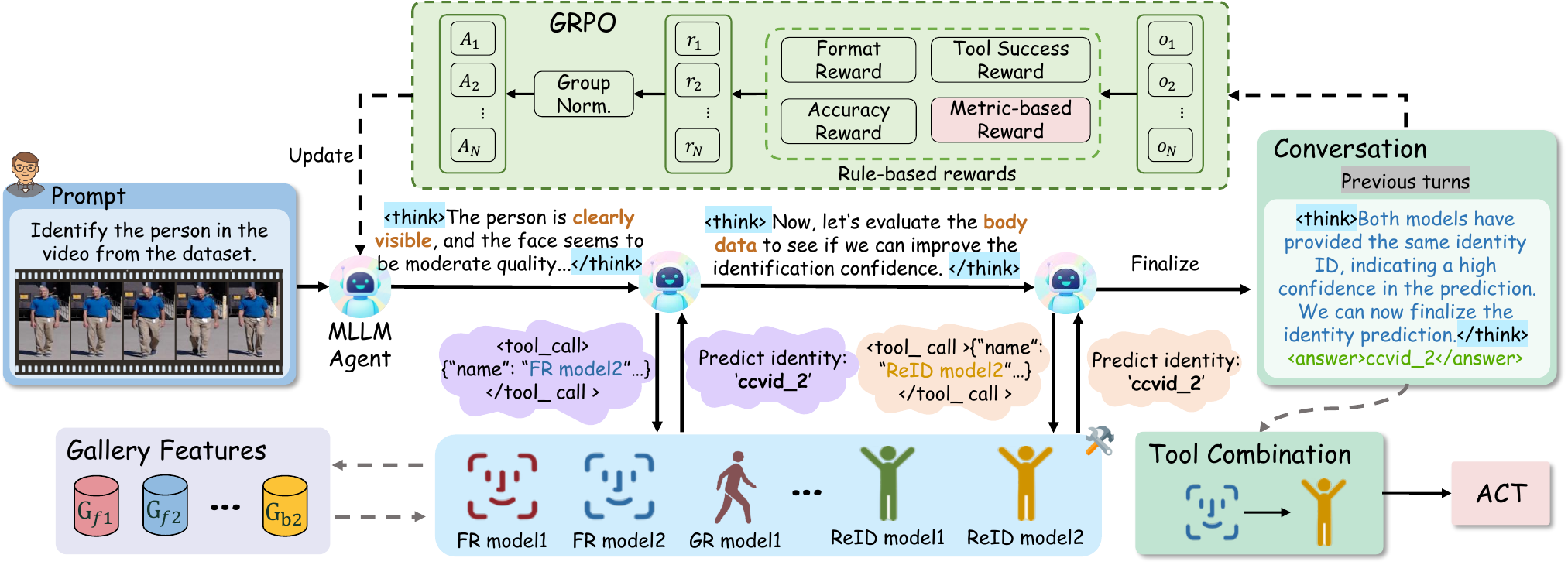}
    \caption{\textbf{Overview of the \ours framework.} Recognition models are wrapped as tools to generate score vectors and predicted identities based on gallery features. The MLLM agent receives multimodal biometric inputs and performs a reasoning-action step through multi-turn, selectively invokes tools, and integrates predictions into a final identity decision and fused score vector. The agent is optimized with reinforcement fine-tuning using rule-based rewards, including the proposed metric-based reward.}
    \label{fig:overview}
\end{figure*}

\subsection{Agents and Reinforcement Learning}

The advent of tool-augmented MLLMs has spurred the development of agentic systems: models capable of planning, reasoning, and interacting with external tools to solve complex tasks. Recent studies~\cite{yao2023react, liu2025arft, chen2025lvagent, zuo20254kagent, xiao2026deepfake} have equipped MLLMs with tools to address diverse challenges such as visual grounding, image super-resolution, and advanced visual understanding. Furthermore, subsequent research has employed RFT with rule-based rewards, such as Group Relative Policy Optimization (GRPO)~\cite{shao2024deepseekmath}, which offers improved generalizability with lower data dependency~\cite{liu2025rft, yu2025perception, shen2025vlm, zhu2026can, xu2025stare}. 
Building upon these advancements, we further propose a metric-based reward to optimize model selections of \ours, effectively capturing data-dependent patterns.

\section{Proposed Method} \label{sec:method}

\paragraph{Problem Formulation.} In biometric recognition, a query (or probe) is a sample sequence that is to be identified (1:N comparisons) or verified by comparison (1:1 comparison) with a gallery of previously enrolled subjects in the system. Let $Q$ denote the set of query videos and $G$ be the gallery. $M$ is a predefined model set $\{m_1, m_2, \dots, m_Z\}$ that is used for feature extraction. For a query $q\in Q$ and a model $m\in M$ (from a predefined model set $M$), $m$ extracts features from $q$ and produces a similarity score vector $\mathbf{s}_{m,q}\in\mathbb{R}^{|G|}$ with respect to each gallery entry $g\in G$. The human recognition pipeline comprises two stages: (i) \emph{sample-level model subset selection} and (ii) \emph{query-based score fusion of selected models}. For each $q$, we seek an optimal subset of models $\mathbf{M}_q\subseteq M$ that maximizes performance. Given $\mathbf{M}_q$, we stack the per-model scores into a score matrix $\mathbf{S}_q\in\mathbb{R}^{|M_q|\times |G|}$ whose $i$-th row is $\mathbf{s}_{m_i,q}$. A score-fusion function $f$ then maps $\mathbf{S}_q$ to a fused score vector $\mathbf{s'}_q=f(\mathbf{S}_q)\in\mathbb{R}^{|G|}$. We detail the model-selection procedure using an agent in Sec.~\ref{subsec:agent}, the supervision signals (reward functions) in Sec.~\ref{subsec:reward_functions}, and the proposed score-fusion method in Sec.~\ref{subsec:act}.

\subsection{Agentic Training Framework} \label{subsec:agent}

Training the agent presents two challenges: (i) the lack of ground-truth dialogue data for Supervised Fine-tuning (SFT), since grid searching optimal model combinations per sample is infeasible; and (ii) the intractability of dataset-level grid search for the best model combination per sample due to exponential growth in the search space with models and samples. We thus employ GRPO, which enables trial-based learning without pre-labeled data. Additionally, we design a metric-based reward function that promotes exploration of diverse model combinations and facilitates implicit learning of the relationship between input characteristics and model performance, as detailed in Sec.~\ref{subsec:reward_functions}.

The overall framework is depicted in Fig.~\ref{fig:overview}. For a given query $q$, the agent first analyzes its content and selects an initial model $a_q \in M_q$. For example, FR model 2 is selected because a clearly visible face is detected. Upon a successful function call, the tool returns $s_{a_q}(q)$ and the predicted label. Since score vectors are not directly interpretable by the agent, the predicted identity label is returned explicitly, while $s_{a_q}(q)$ is retained internally. The agent then decides whether to invoke another model or output its decision. This sequential decision-making process enables the flexible maximization of a long-term objective.~\cite{yao2023react, shinn2023reflexion}. If the agent decides to output its decision, it also provides a reasoning summary that justifies the model selections. Reasoning at each step is necessary to enhance transparency and decision traceability. The whole conversation $o_i$, including the successful model selections, is then used to measure the overall rewards. In GRPO, we sample $N$ rollouts (\ie, responses) for each query and jointly compute their advantages $\{A_1, A_2, \dots, A_N\}$ from the corresponding rewards $\{r_1, r_2, \dots, r_N\}$ to update the agent. Details of GRPO are provided in the supplementary.

\paragraph{Multi-turn Design.}
We adopt a ReAct-style (reason-before-act) multi-turn controller for tool use, rather than a single-shot plan generator~\cite{yao2023react}. This design is motivated by the need to decompose the complex task of multi-tool usage into a sequence of simpler, more manageable decisions. A single-turn approach would require the agent to generate a complete and static execution plan at once, a task plagued by a combinatorial action space and an inability to handle unexpected outcomes. This interleaving of reasoning with actions decomposes multi-tool execution into atomic steps and yields the following benefits: 

\noindent \textbf{i) Simplified Learning:} Reduces the vast action space of generating a full plan to a single decision at each step, making the policy significantly more tractable to learn.

\noindent \textbf{ii) Dynamic Adaptation:} Allows the agent to observe tool outputs and adjust its strategy in real time, enabling error correction and flexible reasoning. These sequential actions also support effective credit assignment during inference.

\subsection{Reward Functions} \label{subsec:reward_functions}

Reward functions play a central role in RFT-based agent training. We design four reward functions: format reward $R_{f}$, tool success reward $R_{tool}$, answer accuracy reward $R_{acc}$, and metric-based reward $R_{mat}$. 

\paragraph{Format Reward.} The agent is trained to produce structured responses that separate reasoning, tool calls, and final answers~\cite{shao2024deepseekmath, liu2025arft, liu2025rft, zhu2026can}. We adapt this to a multi-turn setting where each turn must be a structured format, such as \think\answer. The reward is computed per turn, and the overall reward is averaged across all turns.

\paragraph{Tool Success Reward.} This reward assesses whether the agent's tool calls are executable and yield valid results. Each tool call receives a binary success/failure score, and the reward is defined as the success rate across all tool calls in a trajectory. This incentivizes the agent to produce syntactically correct tool inputs.

\paragraph{Answer Accuracy Reward.} This reward measures the correctness of the agent's final prediction $a_i$ extracted from the answer tag \answer of the response $o_i$ against the ground truth label $y$. Its primary objective is to prioritize factual accuracy over procedural correctness, which enables the agent to implicitly assess model reliability, especially when model predictions are inconsistent:
\begin{equation}
    R_{acc}(a_i, y)=\begin{cases}
                    1, & \text{if~}a_i=y, \\
                    0, & \text{otherwise}. 
                    \end{cases}
\end{equation}

\paragraph{Metric-based Reward.} The metric-based reward guides the agent toward effective and dynamic model selection. A key challenge is to encourage sufficient exploration of diverse model combinations without prior knowledge of their optimality. To address this, the reward is computed based on the performance of the agent-selected model combination on the training set, using the score-fusion method from Sec.~\ref{subsec:act}. The combination is derived from successfully executed conversations per sample.

However, exhaustively searching for the ground-truth optimal model combination for each sample is computationally infeasible. To promote exploration, we construct an augmented model selection $\mathbf{M_Q} \in \mathbb{R}^{|Q|\times|M|}$ based on the model combination $M_{o_i}$ of the response $o_i$, since TAR and FNIR are determined by thresholds selected from \emph{dataset-level} scores. Specifically, a fraction $\gamma \in [0,1]$ of samples in $\mathbf{M_Q}$ retain the same model combination as $M_{o_i}$. For the remaining $(1-\gamma)$, the mask entries are sampled from a Bernoulli distribution to diversify the model combinations explored. The column corresponding to the anchor model is always set to true, ensuring its consistent inclusion.

We apply the \fusion (Anchor-based Confidence Top-k)  (details in Sec.~\ref{subsec:act}) based on $\mathbf{M_Q}$. The performance of the resulting fused scores matrix $\mathbf{S}'_{o_i}$ is then evaluated across the entire training set using four key metrics: True Accept Rate at a False Acceptance Rate (TAR@FAR), mean Average Precision (mAP), Rank-1 accuracy, and False Non-Identity Rate at a specified False Positive Identification Rate (FNIR@FPIR). The final metric-based reward $R_{mat}$ is a composite score formulated to provide a holistic performance signal:
\begin{equation} 
    \mathbf{S}'_{o_i} = \text{ACT}(\mathbf{M_Q}, k),
\end{equation}
\begin{equation} \label{eq:metric_reward}
\begin{split}
    R_{mat}(o_i)= \text{Rank} (\mathbf{S}'_{o_i}) + \text{mAP} (\mathbf{S}'_{o_i}) \\ +\text{TAR} (\mathbf{S}'_{o_i}) - \text{FNIR} (\mathbf{S}'_{o_i}).
\end{split}
\end{equation}

This formulation naturally aligns with both our optimization objective and real-world deployment needs. It rewards the agent for achieving higher accuracy and retrieval performance (TAR, mAP, Rank) while penalizing missed identifications (FNIR), thereby capturing operational performance requirements in a comprehensive manner. The final reward is the sum of $R_{f}$,  $R_{tool}$, $R_{acc}$, and $R_{mat}$.

\begin{figure}[t!]
    \centering
    \includegraphics[width=\linewidth]{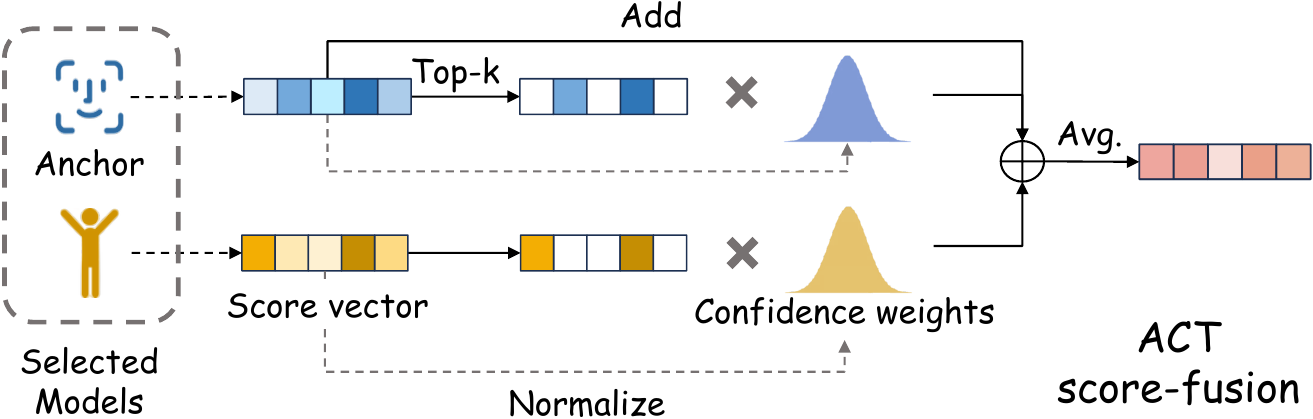}
    \caption{\textbf{Overview of the ACT score-fusion.} Based on tool execution results (\ie, score vectors) and selected model combination, the first selected model serves as the anchor, and ACT produces the final score vector via confidence weighting and top-k filtering.}
    \label{fig:act}
\vspace{-0.5em}
\end{figure}

\subsection{Anchor-based Confidence Top-k Score-fusion} \label{subsec:act}

The proposed \fusion score-fusion approach aims to dynamically and effectively combine scores from multiple models for a given query. The overview is shown in Fig.~\ref{fig:act}. Our approach is built on the principle of leveraging a stable and powerful ``anchor model" $m_a \in \mathbf{M}_q$ to provide a robust score vector, while selectively incorporating normalized, high-confidence scores from the set of models. Let $m_a$ be the first selected model in $\mathbf{M}_q$, and $s_{m,q,g}$ denotes the similarity score between $q$ and $g$ for model $m$.

\begin{figure}
    \centering
    \includegraphics[width=\linewidth]{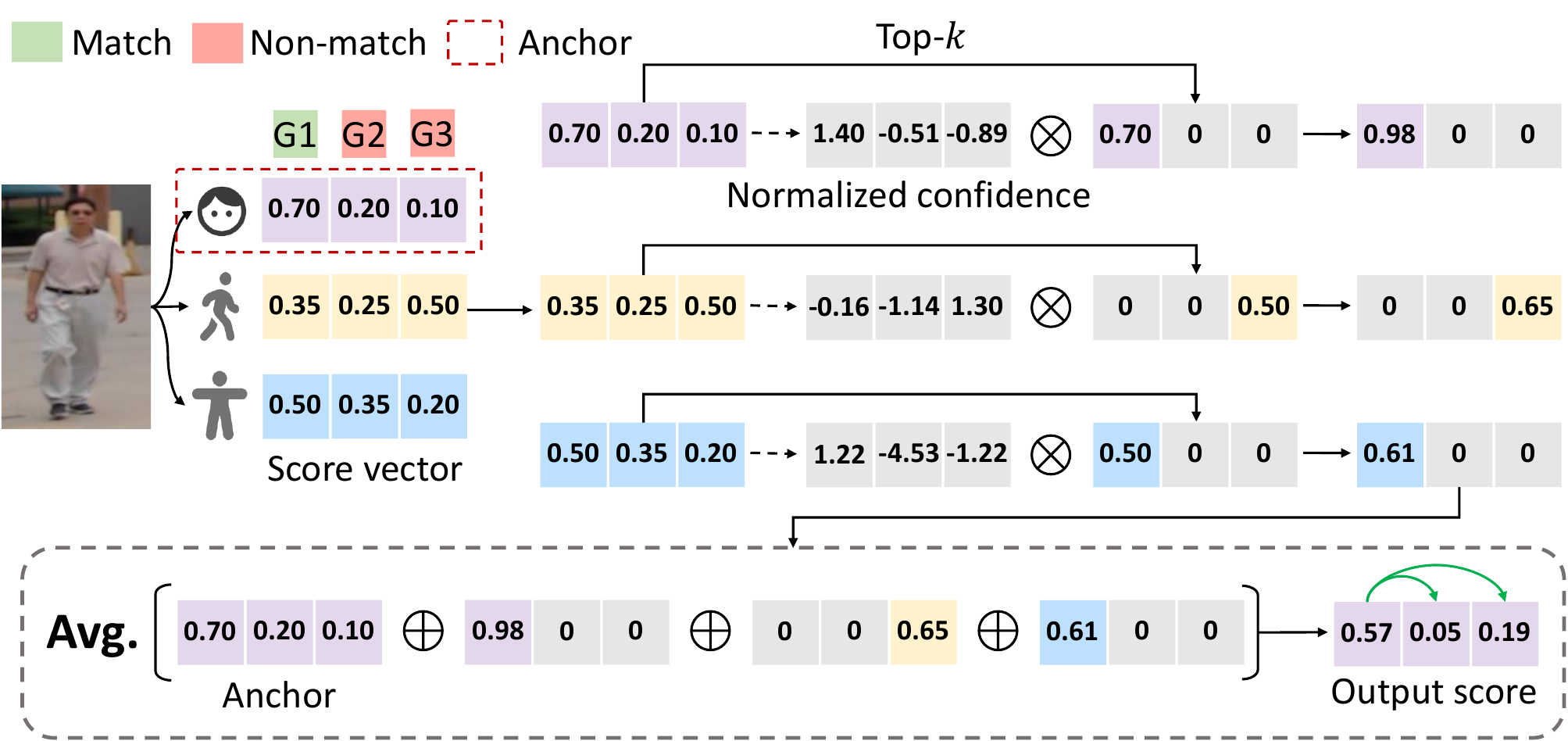}
    \caption{\textbf{A toy example of the proposed ACT score-fusion.} Three models are used with the FR model as the anchor and $k=1$. ACT amplifies the gap between match and non-match scores through confidence-based top-k and anchor weighting, improving verification and open-set search performance.}
    \label{fig:act_example}
\vspace{-0.5em}
\end{figure}

We begin by computing a contribution score $c_{m,q,g}$. This score is designed to leverage the most confident predictions while filtering out potential noise from low-scoring candidates, which is achieved through a Top-k selection process. Formally, for a score $c_{m,q,g}$ in $\mathbf{c}_{m,q} \in \mathbb{R}^{|G|}$, the contribution score is defined as:
\begin{equation}
c_{m,q,g} =
\begin{cases}
z_{m,q,g} \cdot s_{m,q,g} & \text{if } g \in \mathcal{T}_{m,q}, \\
0 & \text{otherwise}.
\end{cases}
\label{eq:contribution_score}
\end{equation}
where $\mathcal{T}_{m,q}$ is the set of indices of the $k$ highest-scoring gallery entries for query $q$ from model $m$. The term $z_{m,q,g}$ represents the Z-score normalized score between $q$ and $g$ by model $m$. This standardization makes the scores from different models with potentially different scales comparable. It is calculated as:
\begin{equation}
z_{m,q,g} = \frac{s_{m,q,g} - \mu_{m,q}}{\sigma_{m,q}},
\end{equation}
where $\mu_{m,q}$ and $\sigma_{m,q}$ are the mean and standard deviation of $\mathbf{s}_{m,q}$. The final fused score vector $\mathbf{s}'_{q}$ is computed via:
\begin{equation}
\mathbf{s}'_{q} = \frac{1}{1 + |\mathbf{M}_q|} \left( \mathbf{s}_{m_a,q} + \sum_{m \in \mathbf{M}_q} \mathbf{c}_{m,q} \right).
\label{eq:final_score}
\end{equation}
The term $(1 + |\mathbf{M}_q|)$ serves as the normalization factor, ensuring a balanced contribution from all models. This strategy balances robust, general performance with specialized, high-confidence insights. The anchor model provides a more substantial contribution by establishing a global ranking structure, as its scores are applied unconditionally to all candidates. In contrast, the selected models provide sparse, localized refinements only for their top-k predictions. Fig.~\ref{fig:act_example} provides a toy example of ACT score-fusion.

\section{Experiments} \label{sec:experiments}

\paragraph{Datasets and Evaluation Protocol.} We evaluate on popular human recognition datasets: CCVID~\cite{gu2022clothes}, MEVID~\cite{davila2023mevid}, and LTCC~\cite{qian2020long}. They comprise multi-view captures and cross-modal biometric data with diverse resolutions and temporal dynamics. Our evaluation protocol follows~\cite{zhu2025quality} for fair comparison. 
Performance is measured via standard metrics in general settings, \ie, Rank-1 accuracy, mean Average Precision (mAP), verification rate (TAR@1.0\%FAR), and open-set search metric (FNIR@1.0\%FPIR), which collectively reflect real-world deployment requirements. 
For open-set search evaluation, we follow the protocol of~\cite{zhu2025quality, su2025open} to construct 10 random subsets of gallery subjects. Each subset, containing  $\sim$$20\%$ of the subjects in the test set, serves as a non-mated list. 
We report the mean and standard deviation for the 10 trials.

\paragraph{Baselines.} Our baselines include the following methods: Min/Max score-fusion~\cite{jain2005score}, Z-score and min-max normalization~\cite{snelick2003multimodal}, RHE~\cite{he2010performance}, FarSight~\cite{liu2024farsight}, and learning-based methods like BSSF~\cite{teng2022optimized}, Weighted-sum~\cite{park2021multi}, AsymA-O1~\cite{herbadji2020combining}, and QME~\cite{zhu2025quality}. We follow ~\cite{zhu2025quality} to construct the biometric model pool (details provided in the Supp.). We also compare with the SoTA multimodal human recognition model SapiensID~\cite{kim2025sapiensid}.

\paragraph{Implementation Details.} Following~\cite{zhu2025quality}, we extract the center embeddings of subjects for training and gallery embeddings for testing by applying average pooling. Considering runtime efficiency in practical scenarios, \ours is based on Qwen2.5-VL-3B~\cite{bai2025qwen2} and fine-tuned via GRPO~\cite{shao2024deepseekmath}. 
For efficiency, we set a turn limit of 4 and train for 200 steps. 
All biometric models are frozen in training. 
For GRPO, we set $N=6$ and a KL coefficient $\beta = 0.04$. 
LoRA~\cite{hu2022lora} is applied with rank $r = 64$ and scaling factor $\alpha = 128$. 
The learning rate follows a linear decay schedule starting from $2\times10^{-5}$. 
The metric-based reward uses $\gamma = 0.8$. 
Based on training-set performance, we set $k = 10$ for CCVID and $k = 40$ for MEVID and LTCC. 
During training, we sample a continuous 4-frame clip from each video for CCVID and MEVID, and 1 frame for LTCC, as it is an image-based dataset.
Training is conducted on 4 H100 GPUs with an effective batch size of 4, and takes nearly 4 hours. During evaluation, we disable sampling in \ours to ensure results are reproducible.

\begin{table}[t!]
\centering
\tabcolsep=0.1cm
 \resizebox{\linewidth}{!}{
        \begin{tabular}{ccccccc}
            \toprule
            \textit{Method} &\textit{Comb.}  & Rank1$\uparrow$ & mAP$\uparrow$ & TAR$\uparrow$ & FNIR$\downarrow$  \\ \midrule
            \textit{AdaFace}$^*$~\cite{kim2022adaface} & \textcolor{Peach}{$\blacklozenge$} & $94.0$ & $87.9$ & $75.7$ & $13.0\pm3.5$ \\
            \textit{CAL}~\cite{gu2022clothes}  & \textcolor{Cyan}{$\spadesuit$} & $81.4$ & $74.7$ & $66.3$ & $52.8\pm13.3$ \\ 
            \textit{BigGait}$^*$~\cite{ye2024biggait} & \textcolor{JungleGreen}{$\clubsuit$} & $76.7$ & $61.0$ & $49.7$ & $71.1\pm6.1$ \\ 
            \textit{SapiensID}~\cite{kim2025sapiensid} & \textcolor{Purple}{\ding{108}} & $92.6$ & $77.8$ & - & - \\           
            \midrule
            \textit{Min-Fusion}~\cite{jain2005score} & \multirow{12}{*}{\textcolor{Peach}{$\blacklozenge$} \textcolor{Cyan}{$\spadesuit$} \textcolor{JungleGreen}{$\clubsuit$}} & $87.1$ & $79.2$ & $62.4$ & $48.5 \pm 8.7$ \\
            \textit{Max-Fusion}~\cite{jain2005score} & & $89.9$ & $89.3$ & $73.4$ & $23.0\pm10.1$ \\
            \textit{Z-score}~\cite{snelick2003multimodal} & & $92.2$ & $90.6$ & $73.9$ & $15.1\pm1.5$ \\
            \textit{Min-max}~\cite{snelick2003multimodal} & & $91.8$ & $90.9$ & $73.9$ & $15.4\pm2.5$ \\
            \textit{Weighted-sum}~\cite{park2021multi} & & $91.7$ & $90.6$ & $73.6$ & $15.4\pm1.8$  \\
            \textit{Asym-AO1}~\cite{herbadji2020combining} & & $92.3$ & $90.0$ & $74.0$ & $15.9\pm1.7$ \\
            \textit{BSSF}~\cite{teng2022optimized} & & $91.8$ & $91.1$ & $73.9$ & $14.1\pm1.3$ \\             
            \textit{FarSight}~\cite{liu2024farsight} & & $92.0$ & $91.2$ & $73.9$ & $13.9\pm1.1$ \\
            \textit{QME}~\cite{zhu2025quality} & & $\mathbf{94.1}$ & $90.8$ & $76.2$ & $12.3\pm1.4$ \\
            \rowcolor{sky!30}\textit{\ours (DA)} & & {$92.8$} & $\underline{92.2}$ & $\underline{85.8}$ & $\underline{10.5\pm1.5}$ \\ 
            \rowcolor{sky!30}\textit{\ours (CoT)} & & \underline{$93.4$} & $\mathbf{92.6}$ & $\mathbf{85.9}$ & $\mathbf{10.1\pm1.5}$ \\ 
            \bottomrule
        \end{tabular}
    }
\caption{\textbf{Performance on CCVID.} [Keys: \textbf{Best} and \underline{second best} performance; \textit{Comb.}: model combination; $^*$: zero-shot performance; \textcolor{Peach}{$\blacklozenge$}: AdaFace for face modality; \textcolor{JungleGreen}{$\clubsuit$}: BigGait for gait modality; \textcolor{Cyan}{$\spadesuit$}: CAL of body modality; \textcolor{Purple}{\ding{108}}: SapiensID for face and body modalities; TAR: TAR@1\%FAR; FNIR: FNIR@1\%FPIR.]}
\label{tab:ccvid_performance}
\vspace{-0.5em}
\end{table}

\begin{table}[t!]
\centering
\tabcolsep=0.1cm
 \resizebox{\linewidth}{!}{
        \begin{tabular}{ccccccc}
            \toprule
            \textit{Method} &\textit{Comb.}  & Rank1$\uparrow$ & mAP$\uparrow$ & TAR$\uparrow$ & FNIR$\downarrow$  \\ \midrule
            \textit{AdaFace}$^*$~\cite{kim2022adaface} & \textcolor{Peach}{$\blacklozenge$} & $18.5$ & $5.9$ & $2.4$ & $99.8\pm0.2$ \\
            \textit{CAL}~\cite{gu2022clothes} & \textcolor{Cyan}{$\spadesuit$} & $74.4$ & $40.6$ & $36.7$ & $59.7\pm7.3$ \\ 
            \textit{AIM}~\cite{yang2023good} & \textcolor{Cyan}{$\blacksquare$} & $74.8$ & $40.9$ & $37.0$ & $66.2\pm7.5$ \\ 
            \textit{SapiensID}~\cite{kim2025sapiensid} & \textcolor{Purple}{\ding{108}} & $72.0$ & $34.6$ & - & - \\      
            \midrule
            \textit{Min-Fusion}~\cite{jain2005score} & \multirow{10}{*}{\textcolor{Peach}{$\blacklozenge$} \textcolor{Cyan}{$\spadesuit$} \textcolor{Cyan}{$\blacksquare$}} & $38.1$ & $13.5$ & $12.4$ & $81.9\pm6.0$ \\
            \textit{Max-Fusion}~\cite{jain2005score}  & & $62.5$ & $33.3$ & $16.8$ & $94.8\pm4.7$ \\
            \textit{Z-score}~\cite{snelick2003multimodal} & & $73.0$ & $37.5$ & $30.4$ & {$68.7\pm9.2$} \\
            \textit{Min-max}~\cite{snelick2003multimodal}  & & $73.2$ & $38.1$ & $31.9$ & $75.1\pm9.2$ \\
            \textit{Weighted-sum}~\cite{park2021multi} & & $73.2$ & $37.8$ & $31.3$ & $72.4\pm8.6$ \\ 
            \textit{Asym-AO1}~\cite{herbadji2020combining} & & $71.2$ & $32.9$ & $19.1$ & $76.3\pm8.9$ &  \\
            \textit{BSSF}~\cite{teng2022optimized} & & ${73.5}$ & {$39.1$} & {$34.2$} & $68.9\pm8.5$ \\                 
            \textit{FarSight}~\cite{liu2024farsight}  & & $73.2$ & $37.8$ & $31.3$ & $72.4\pm8.6$ \\
            \textit{QME}~\cite{zhu2025quality} & & ${73.8}$ & ${39.6}$ & ${35.0}$ & ${64.3\pm8.0}$ \\
            \rowcolor{sky!30}\textit{\ours (DA)} & & $\mathbf{75.5}$ & $\mathbf{41.0}$ & $\underline{36.5}$ & $\underline{50.3\pm9.0}$ \\ 
            \rowcolor{sky!30}\textit{\ours (CoT)} & & $\mathbf{75.5}$ & $\mathbf{41.0}$ & $\mathbf{37.0}$ & $\mathbf{50.0\pm8.5}$ \\ 
            \bottomrule
        \end{tabular}
    }
\caption{\textbf{Performance on LTCC.} [Keys: \textbf{Best} and \underline{second best} performance; \textit{Comb.}: model combination; $^*$: zero-shot performance; \textcolor{Peach}{$\blacklozenge$}: AdaFace for face modality; \textcolor{Cyan}{$\spadesuit$}: CAL of body modality; \textcolor{Cyan}{$\blacksquare$}: AIM for body modality; \textcolor{Purple}{\ding{108}}: SapiensID for face and body modalities; TAR: TAR@1\%FAR; FNIR: FNIR@1\%FPIR.]}
\label{tab:ltcc_performance}
\vspace{-1em}
\end{table}

\begin{table}[t!]
\centering
\tabcolsep=0.1cm
 \resizebox{\linewidth}{!}{
        \begin{tabular}{ccccccc}
            \toprule
            \textit{Method} &\textit{Comb.}  & Rank1$\uparrow$ & mAP$\uparrow$ & TAR$\uparrow$ & FNIR$\downarrow$  \\ \midrule
            \textit{AdaFace}$^*$~\cite{kim2022adaface} & \textcolor{Peach}{$\blacklozenge$} & $25.0$ & $8.1$ & $5.4$ & $98.8\pm1.2$ \\
            \textit{CAL}~\cite{gu2022clothes} & \textcolor{Cyan}{$\spadesuit$} & $52.5$ & $27.1$ & $34.7$ & $67.8\pm7.3$ \\ 
            \textit{AGRL}~\cite{wu2020adaptive} & \textcolor{Cyan}{$\blacksquare$} & $51.9$ & $25.5$ & $30.7$ & $69.4\pm8.9$ \\ 
            \midrule
            \textit{Min-Fusion}~\cite{jain2005score} & \multirow{11}{*}{\textcolor{Peach}{$\blacklozenge$} \textcolor{Cyan}{$\spadesuit$} \textcolor{Cyan}{$\blacksquare$}} & $46.8$ & $21.2$ & $28.0$ & $70.4\pm8.0$ \\
            \textit{Max-Fusion}~\cite{jain2005score}  & & $33.2$ & $14.9$ & $8.3$ & $97.4\pm1.6$ \\
            \textit{Z-score}~\cite{snelick2003multimodal} & & $54.1$ & $27.4$ & $30.7$ & $66.5\pm7.0$ \\
            \textit{Min-max}~\cite{snelick2003multimodal}  & & $52.8$ & $24.7$ & $25.0$ & $71.3\pm6.1$ \\
            \textit{Weighted-sum}~\cite{park2021multi} & & $54.1$ & $27.3$ & $30.3$ & $66.3\pm7.0$ \\            
            \textit{Asym-AO1}~\cite{herbadji2020combining} & & $52.5$ & $22.9$ & $23.6$ & $71.7\pm5.8$ \\
            \textit{BSSF}~\cite{teng2022optimized} & & $53.5$ & $27.4$ & $30.5$ & $65.9\pm7.2$ \\          
            \textit{FarSight}~\cite{liu2024farsight}  & & $53.8$ & $25.4$ & $26.6$ & $69.8\pm6.4$ \\
            \textit{QME}~\cite{zhu2025quality} & & $\mathbf{55.7}$ & $\underline{28.2}$ & $32.9$ & $64.6\pm8.2$ \\
            \rowcolor{sky!30}\textit{\ours (DA)} & & ${52.5}$ & $\mathbf{28.7}$ & $\underline{34.8}$ & $\underline{60.8\pm7.3}$ \\
            \rowcolor{sky!30}\textit{\ours (CoT)} & & $\underline{54.7}$ & $\mathbf{28.7}$ & $\mathbf{34.9}$ & $\mathbf{58.6\pm7.4}$ \\
            \bottomrule
        \end{tabular}
    }
\caption{\textbf{Performance on MEVID.} [Keys: \textbf{Best} and \underline{second best} performance; \textit{Comb.}: model combination; $^*$: zero-shot performance; \textcolor{Peach}{$\blacklozenge$}: AdaFace for face modality; \textcolor{Cyan}{$\spadesuit$}: CAL of body modality; \textcolor{Cyan}{$\blacksquare$}: AGRL for body modality; TAR: TAR@1\%FAR; FNIR: FNIR@1\%FPIR.]}
\label{tab:mevid_performance}
\vspace{-0.3em}
\end{table}

\begin{table}[t!]
\centering
\tabcolsep=0.1cm
 \resizebox{\linewidth}{!}{
        \begin{tabular}{cccccc}
            \toprule
            \textit{Method} & Rank1$\uparrow$ & mAP$\uparrow$ & TAR$\uparrow$ & FNIR$\downarrow$  \\ \midrule
            \rowcolor{gray!30} \multicolumn{6}{c}{\textit{CCVID $\rightarrow$ LTCC (Zero-shot)}} \\
            \textit{\ours (\fusion)} & $60.4$ & $11.9$ & $7.7$ & $60.3\pm8.6$ &  \\ 
            \textit{\ours (FarSight)} & $68.2$ & $31.7$ & $17.0$ & $81.8\pm9.6$ &  \\ 
            \rowcolor{gray!30} \multicolumn{6}{c}{\textit{CCVID $\rightarrow$ LTCC (10-shot)}} \\            
            \textit{\ours (\fusion)} & $73.6$ & $39.8$ & $34.8$ & $53.5\pm8.5$ \\  
            \rowcolor{gray!30} \multicolumn{6}{c}{\textit{MEVID $\rightarrow$ LTCC (Zero-shot)}} \\
            \textit{\ours (FarSight)} & $75.3$ & $42.3$ & $37.7$ & $59.3\pm9.5$ &  \\ 
            \textit{\ours (\fusion)} & $75.3$ & $41.1$ & $36.1$ & $50.1\pm8.4$ &  \\ 
            
            \bottomrule
        \end{tabular}
    }
\caption{\textbf{Cross-domain Performance on LTCC.} \ours is trained on CCVID with its model combination (Tab.~\ref{tab:ccvid_performance}) and evaluated on LTCC using LTCC’s model combination  (Tab.~\ref{tab:ltcc_performance}). [Keys: TAR: TAR@1\%FAR; FNIR: FNIR@1\%FPIR.] }
\label{tab:ltcc_cross_performance}
\vspace{-0.5em}
\end{table}

\subsection{Experimental Results}

\paragraph{In-domain Evaluation.} As shown in Tab.~\ref{tab:ccvid_performance}, ~\ref{tab:ltcc_performance}, and~\ref{tab:mevid_performance}, the upper section of each table lists the unimodal models, while the lower section reports the performance of fusion methods. Statistical baselines (\eg, Z-score, FarSight) often fail to surpass the strongest single model across all metrics, while learning-based methods (\eg, QME) achieve better gains and occasionally outperform the best unimodal model, but remain limited when fused models are not complementary (\eg, LTCC). By contrast, \ours consistently achieves superior performance on all three datasets, outperforming both unimodal and fusion baselines on most metrics. Across the tables, FNIR benefits the most from fusion, while Rank-1 improves the least. This is because FNIR is particularly sensitive to outliers, whereas Rank-1 is largely determined by the strongest modality and only marginally affected by fusion. Through top-$k$ selection and confidence weighting, our method effectively bounds the increase in non-match scores. On CCVID, the largest TAR gain arises because \ours identifies high-quality facial inputs and primarily anchors on the FR model, which excels at distinguishing matches from non-matches. On LTCC, FNIR reduction is most pronounced due to the proposed ACT score-fusion strategy (see ablation in Tab.~\ref{tab:ablation_score_fusion_method}).

\paragraph{Efficiency in Real-world Scenarios.} We introduce two inference modes: \da (DA) and Chain-of-Thought (CoT). DA avoids explicit reasoning, reducing average inference time from 2.81s to 1.03s on an H100 GPU, while maintaining competitive performance relative to QME (0.67s). CoT offers interpretable results, but at the cost of higher latency. Users can flexibly choose between the two to balance efficiency, interpretability, and accuracy.

\paragraph{Cross-domain Evaluation.} In real-world deployments, agents may encounter unseen environments with different tool combinations, making cross-domain evaluation critical for assessing transferability. We evaluate \ours by training on CCVID and testing on LTCC (Tab.~\ref{tab:ltcc_cross_performance}). In the zero-shot setting, despite the changed tool pool, \ours generalizes well and successfully executes tool calls without additional training. However, in CCVID $\rightarrow$ LTCC, where face quality degrades, \ours experiences a performance drop due to (1) the unknown robustness of newly introduced models or weights on the target dataset. Inappropriate anchor selection may amplify errors and degrade overall performance. (2) the domain gap in data distribution. With only 10 samples per subject (\ie, $8\%$ of the training data) and 50 training steps, \ours quickly adapts and achieves performance comparable to full in-domain training. Under MEVID $\rightarrow$ LTCC, the zero-shot setting attains similar performance, further demonstrating the robustness of \ours.

\begin{figure}[t!]
    \centering
    \includegraphics[width=.9\linewidth]{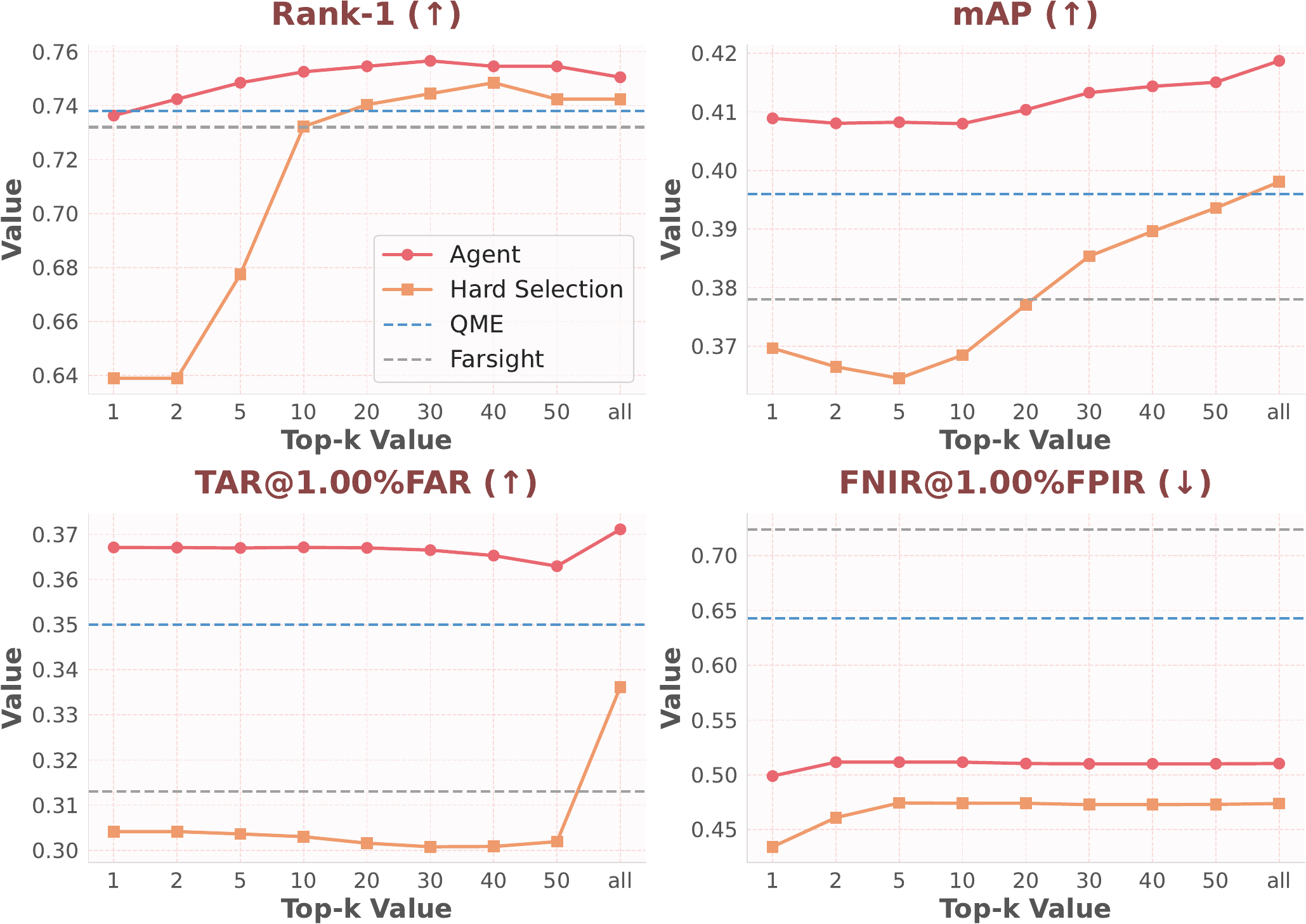}
    \caption{\textbf{Performance comparison on LTCC in four metrics.} \ours consistently outperforms baselines, including the hard selection (\ie, using all models), which highlights the effectiveness of dynamic model selection. 
}   
    \label{fig:ablation_topk}
\vspace{-0.5em}
\end{figure}

\subsection{Ablation Studies} \label{subsec:ablation_studies}

\paragraph{Agent Selection \vs Hard Selection.}
We compare dynamic model selection by the agent (red line in Fig.~\ref{fig:ablation_topk}) with hard selection (orange line), which uses all available models. Since hard selection lacks anchor information, we use the averaged scores as a surrogate anchor, whereas in the agent setting, the first selected model serves as the anchor. Our method yields notable improvements across all metrics over previous SoTA approaches, underscoring the efficacy of the proposed agent and the \fusion algorithm. It also consistently outperforms hard selection in Rank-1, mAP, and TAR, highlighting the critical role of adaptive model selection. The significantly lower performance of hard selection relative to prior SoTA confirms that dynamic model selection is the key driver of performance gains.

\paragraph{Top-K Values.}
We investigate the impact of top-$k$ values in \fusion.
Fig.~\ref{fig:ablation_topk} indicates that the choice of $k$ affects different metrics distinctively: larger $k$ values generally improve Rank-1, mAP, and TAR@FAR, while FNIR remains relatively stable. Compared with hard selection, which exhibits large fluctuations, especially at small $k$, the proposed agent maintains both higher performance and greater stability across a wide range of $k$ values. 
This shows that the agent not only adapts to sample-level variations by selectively exploiting complementary models but also avoids the noise introduced by averaging redundant scores. The robustness of \fusion under varying $k$ highlights its effectiveness in supporting adaptive model selection, ensuring consistent improvements across diverse evaluation settings.
 
\paragraph{Alternative Score-fusion Methods.}
Tab.~\ref{tab:ablation_score_fusion_method} compares the proposed \fusion against alternative statistical score-fusion methods using the same agent-selected model combinations. 
Even with standard fusion techniques such as Z-score and FarSight, our approach already surpasses the QME, highlighting the decisive role of dynamic model selection. Interestingly, Z-score and FarSight achieve nearly identical results, suggesting that the choice of statistical fusion has a limited impact once adaptive selection is applied. In contrast, the proposed \fusion yields further consistent gains, with a substantial reduction in FNIR (down to 51.0). This demonstrates that beyond dynamic selection, robust score integration is crucial for handling challenging open-set search scenarios, making \fusion more reliable and generalizable across diverse evaluation metrics.

\begin{table}[t!]
\centering
\tabcolsep=0.1cm
\resizebox{0.9\linewidth}{!}{
    \begin{tabular}{l|cccc}
    \toprule
        Method &  Rank1 & mAP & TAR@1\%FAR & FNIR@1\%FPIR   \\
        \midrule
        \textit{QME}~\cite{zhu2025quality} & $73.8$ & $39.6$ & $35.0$ & $64.3\pm8.0$ \\
        \midrule
        \textit{Z-score} & $74.8$ & $\mathbf{41.7}$ & $37.1$ & $63.7 \pm 9.5$ \\
        \textit{FarSight} & $74.8$ & $\mathbf{41.7}$ & $\mathbf{37.2}$ & $62.5 \pm 9.7$ \\
        \bf \textit{\fusion (Ours)} & $\mathbf{75.5}$ & $41.4$ & $36.5$ & $\mathbf{51.0 \pm 9.4}$ \\ 
    \bottomrule
    \end{tabular}
    }%
    \caption{\textbf{Comparison of score-fusion methods with agent selection on LTCC.} All fusion methods combined with agent-based selection outperform QME, confirming the value of dynamic model selection. The proposed \fusion yields the best overall performance, particularly in open-set search.}
    \label{tab:ablation_score_fusion_method}
\vspace{-0.8em}
\end{table}

\begin{figure}[t!]
    \centering
    \includegraphics[width=\linewidth]{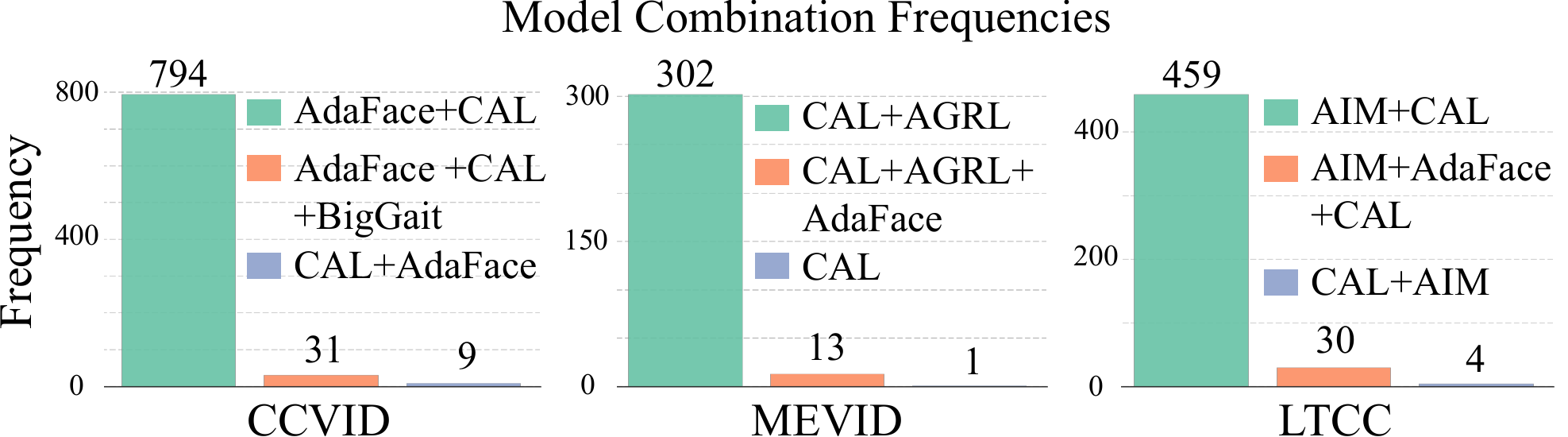}
    \caption{\textbf{Statistics of model selection.} The first model is the anchor model for each category. The combination distribution reveals the characteristics of each dataset and which model combination provides robust performance.}
    \label{fig:ablation_model_selection}
\vspace{-0.8em}
\end{figure}

\paragraph{Statistics of Dynamic Model Selection.}
Fig.~\ref{fig:ablation_model_selection} presents the frequency of model and anchor selections per dataset. On CCVID, where faces are often clearly visible, the agent frequently selects AdaFace as the anchor and consistently combines it with CAL, while BigGait is selected less often, suggesting limited complementary value for combination. In contrast, LTCC and MEVID—collected under surveillance conditions with multi-view and low-quality faces—lead the agent to rely on ReID models.

\begin{table}[t!]
\centering
\tabcolsep=0.1cm
\resizebox{0.9\linewidth}{!}{
    \begin{tabular}{l|cccc}
    \toprule
        Comb. & Rank1 & mAP & TAR@1\%FAR & FNIR@1\%FPIR \\
        \midrule
        \textit{AIM+CAL} & $74.8$ & $41.0$ & $36.1$ & $53.4 \pm 9.2$ \\
        \textit{\ours} & $75.5$ & $41.4$ & $36.5$ & $51.0 \pm 9.4$ \\ 
    \bottomrule
    \end{tabular}
    }
\caption{\textbf{Ablation on model combination and anchor model selection on LTCC.} The results show that \ours closely aligns with the grid-searched results but achieves slightly better performance due to sample-level diversity.}
\label{tab:ablation_grid_search}
\vspace{-1em}
\end{table}

\paragraph{Dataset-level Model Selection.} 
Since exhaustive sample-level grid search is computationally prohibitive, we conduct grid search at the dataset level to determine the optimal model combination and anchor model. As shown in Tab.~\ref{tab:ablation_grid_search}, AIM+CAL is the best model combination using \fusion for the grid search, which is highly consistent with the combination chosen by our agent. Nevertheless, the agent achieves superior performance, attributed to its ability to exploit sample-level diversity during model selection.

\begin{figure}
    \centering
    \includegraphics[width=0.9\linewidth]{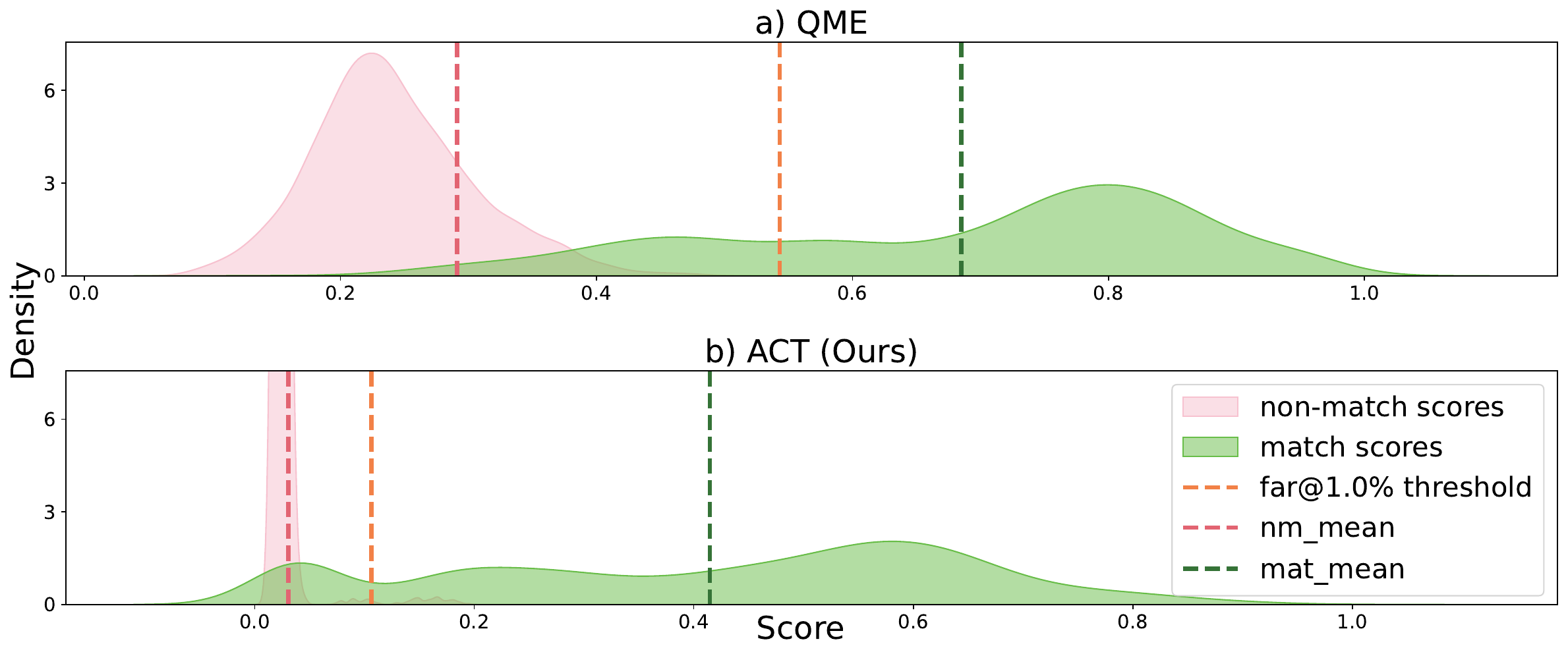}
    \caption{\textbf{Comparison of score distribution on CCVID.} [Keys: nm\_mean = mean value of non-match scores; mat\_mean = mean value of match scores.]}
    \label{fig:qme_ours_score_distribution}
\vspace{-1.5em}
\end{figure}

\begin{figure}[t!]
\vspace{-1em}
    \centering
    \includegraphics[width=0.88\linewidth]{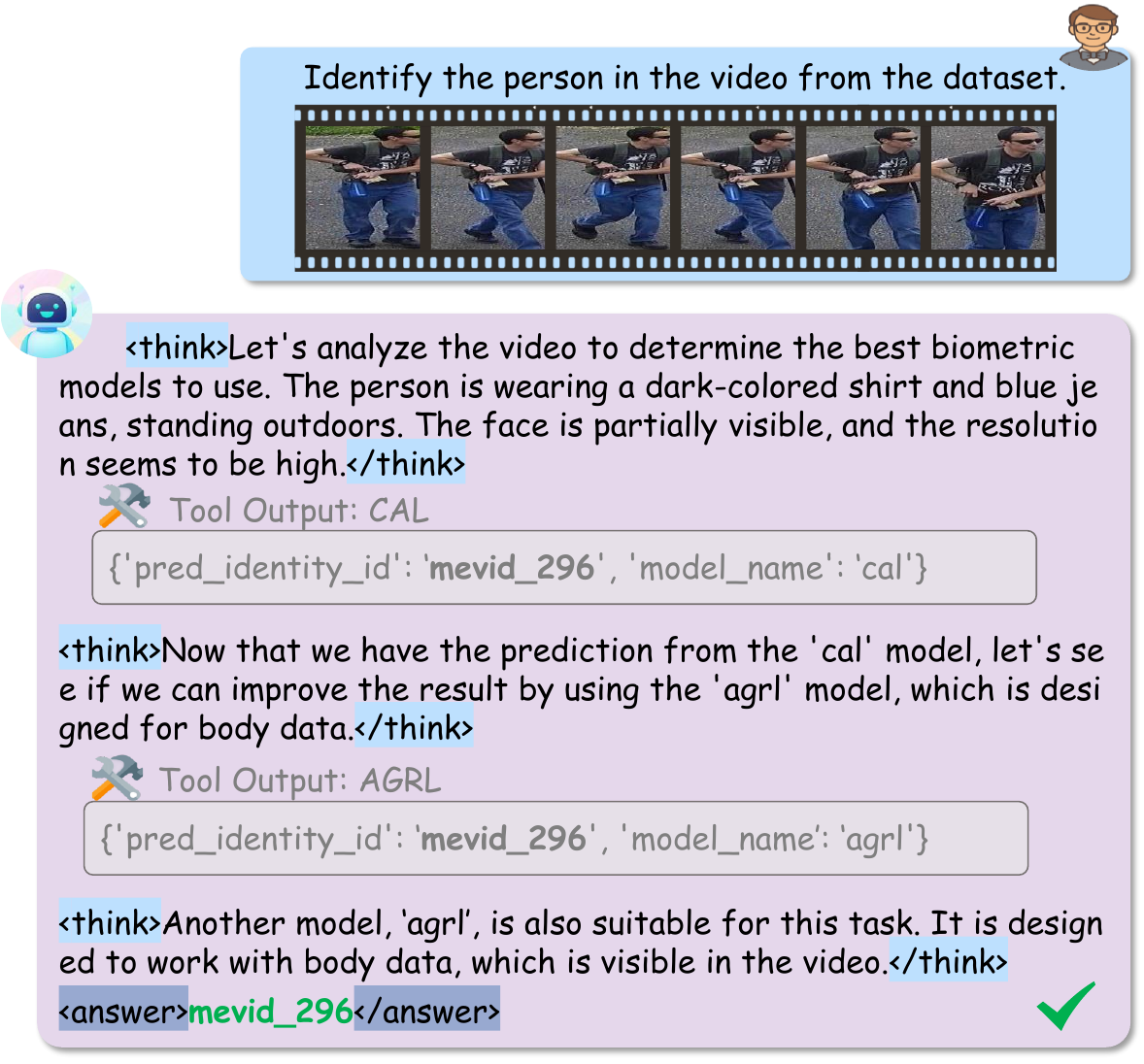}
    \caption{\textbf{CoT of \ours on MEVID.} \ours performs interpretable reasoning and dynamically selects the suitable tool combination for each test sample.}
    \label{fig:visualization_conv}
\vspace{-1em}
\end{figure}

\subsection{Qualitative Results}

\paragraph{Score Distribution.} Fig.~\ref{fig:qme_ours_score_distribution} plots the distribution of non-match (imposter) scores, match (genuine) scores, and the threshold for FAR=1\% for QME and \ours. For clarity, we align the y-axis and normalize the scores. Our method demonstrates a markedly larger margin between the match scores and the FAR threshold, while effectively suppressing non-match scores near zero. This improvement stems from two key mechanisms: (1) by summing only the top-$k$ confident scores, we amplify the contribution of true matches and widen the separation from non-matches, which are predominantly excluded; and (2) our dynamic model selection provides a more robust set of models per query, leading to higher-quality scores for fusion.

\paragraph{CoT for Model Selection.}
Fig.~\ref{fig:visualization_conv} illustrates the CoT mode of \ours, where the agent interprets the input and dynamically selects suitable models for each query sample. By leveraging the learned reliability and complementarity of different models, \ours reasons over the query’s characteristics to determine the optimal model combination. This CoT process enhances interpretability by explicitly revealing the agent’s decision path and selection rationale.
\section{Conclusion}

We propose \ours, a novel agentic framework for human recognition with dynamic model selection. We introduce an Anchor-based Confidence Top-k score-fusion method (\fusion) for sample-dependent and adaptive model integration. Multiple reward functions are designed to guide the agent's tool usage and exploration of diverse model combinations. Extensive experiments and analyses validate the effectiveness of \ours and \fusion, highlighting the benefit of query-wise model selection and fusion. Our approach provides a scalable and extensible solution for multi-modal and multi-model tasks, with potential applicability to broader vision tasks~\cite{yuan2025survey, zhang2026chain, kumar2025charm3r, zhang2026towards, guo2025rethinking, fu2025learning, shen2025fine, liu2026palm}. We hope this work inspires further research on agentic AI.

\paragraph{Acknowledgments}

This research is based upon work supported in part by the Office of the Director of National Intelligence (ODNI), Intelligence Advanced Research Projects Activity (IARPA), via 2022-21102100004. The views and conclusions contained herein are those of the authors and should not be interpreted as necessarily representing the official policies, either expressed or implied, of ODNI, IARPA, or the U.S. Government. The U.S. Government is authorized to reproduce and distribute reprints for governmental purposes notwithstanding any copyright annotation therein.

{\small
\bibliographystyle{ieee_fullname}
\bibliography{egbib}
}

\clearpage
\clearpage
\setcounter{page}{1}
\maketitlesupplementary

\section{Additional Methodology}

\subsection{Group Relative Policy Optimization (GRPO)}

In contrast to RL algorithms such as Proximal Policy Optimization (PPO)~\citep{schulman2017proximal}— which rely on a critic model to assess policy performance, GRPO eliminates the need for the critic model by directly comparing groups of candidate responses. As shown in Fig.~\ref{fig:overview}, for a given input $x$, GRPO requires the model to sample $N$ diverse responses $\{o_1, o_2, \dots, o_N\}$ from the curren t model $\pi_{\theta}$ and obtains overall rewards $\{r_1, r_2, \dots, r_N\}$ for $o_i$ based on the reward function $R(x, o_i)$. In our case, it can be formatted as:
\begin{equation}
\begin{split}
    R(x, o_i) = w_{f} R_{f}(o_i) + w_{tool} R_{tool}(o_i) \\ +  w_{acc} R_{acc}(a_i, y) + w_{mat} R_{mat}(o_i),
\end{split}
\end{equation}
where $w_{f}$, $w_{tool}$, $w_{acc}$, and $w_{mat}$ are the reward weights for format reward $R_{f}$, tool success reward $R_{tool}$, answer accuracy reward $R_{acc}$, and metric-based reward $R_{mat}$, respectively. GRPO assesses the relative quality by normalizing $r_i$ using the mean and standard deviation of the group reward:
\begin{equation}
    A_i=\frac{r_i-\mathrm{mean}(\{r_1,\ldots,r_G\})}{\mathrm{std}(\{r_1,\ldots,r_G\})},
\end{equation}
where $A_i$ denotes the advantage of the $i$-th response. With the group normalization, GRPO encourages the model to sample preferred answers with a higher reward. The model is updated via:
\begin{equation}
\begin{split}
    J_{\mathrm{GRPO}}(\theta) = \mathbb{E}_{q \sim P(Q), \{o_i\}_{i=1}^G \sim \pi_{\theta_{\mathrm{old}}}(O|q)} \\
    \Bigg[ \frac{1}{G} \sum_{i=1}^G \min \Bigg( \frac{\pi_{\theta}(o_i \mid q)}{\pi_{\theta_{\mathrm{old}}}(o_i \mid q)} A_i, \\ \mathrm{clip} \left( \frac{\pi_{\theta}(o_i \mid q)}{\pi_{\theta_{\mathrm{old}}}(o_i \mid q)}, 1 - \varepsilon, 1 + \varepsilon \right) A_i \Bigg) \\ - \beta D_{\mathrm{KL}} \big( \pi_{\theta} \parallel \pi_{\mathrm{ref}} \big) \Bigg],
\end{split}
\end{equation}

where $\varepsilon$ and $\beta$ are the GRPO clipping hyperparameters and the coefficient weight for controlling the Kullback–Leibler (KL) penalty~\citep{schulman2017proximal}, respectively. $\pi_\mathrm{ref}$ is the reference model.

\begin{align}
\frac{1}{1+\mathit{Euc}(q, g)}.
\label{eq:euc_scores}
\end{align} 

\paragraph{Prompts.}
We provide the system prompt used in the agent training and inference in Fig.~\ref{fig:system_prompt}. The tool schema is the function documentation with input and output formats and meaning. Model type dict is the modality type of each biometric model. 

\begin{figure*}
    \begin{tcolorbox}[size=small, parbox=false,colback=fullgreen!10, colframe=fullgreen!50, title=Judge Prompt, coltitle=black]
    \# Role \& Objective
    
    You are an expert-level biometric analysis agent. Your primary mission is to achieve the highest possible identification performance by strategically analyzing input images/videos and selecting the optimal combination of biometric models. 
    Prioritize the model you think is the most suitable. Do not select the same model more than once. Your final answer should be a fused identity prediction based on the evidence from your chosen models.
    
    \# Loop
    
    Work step-by-step. Each turn you must output exactly TWO blocks—first $<$think$>$, then ONE action: $<$tool\_call$>$ or $<$answer$>$. Wait for $<$tool\_result$>$ before the next turn.
    
    \# Strict Output Format (no extra text, no markdown)
    
    1) $<$think$>$...$<$/think$>$$<$tool\_call$>$\{JSON\}$<$/tool\_call$>$
    
    2) $<$think$>$...$<$/think$>$$<$answer$>$...$<$/answer$>$
    
    \# Tag Rules
    
    - $<$think$>$ (required, first): Briefly describe what you get, and explain the current decision.  
      
    \hspace{1em} • If calling a tool: you MUST first analyze the input video's characteristics. Consider factors like: Is the face clearly visible? Is the subject close to the camera with high resolution, or far away and low-resolution? etc.
      
    \hspace{1em} • If answering: summarize tools results, key evidence, and your final prediction.
    
    - $<$tool\_call$>$: JSON with exactly two keys, "name" and "parameters". You can call ONLY ONE tool per turn. 
    
    - $<$answer$>$: Identity: The ID of the recognized person.
    
    \# Tools 
    
    You may call one or more functions to assist with the user query. 
    
    You are provided with function signatures within $<$tools$>$$<$/tools$>$ XML tags: $<$tools$>$ \{TOOL\_SCHEMA\} $<$/tools$>$ 
    
    For each function call, return a json format object with function name and arguments within $<$tool\_call$>$$<$/tool\_call$>$ XML tags: 
    $<$tool\_call$>$ \{\{"name": $<$function-name$>$, "parameters": $<$args-json-object$>$\}\} $<$/tool\_call$>$. Only call declared tools.
    
    \# Model Type
    
    You have access to a suite of specialized models. Your key challenge is to understand when to use them for maximum impact: \{MODEL\_TYPE\_DICT\}
    
    \# Stopping Condition
    
    End with $<$answer$>$ when evidence is sufficient. Never invent tool outputs or identities.
    
    \end{tcolorbox}
    \caption{System prompt for \ours.}
    \label{fig:system_prompt}
\end{figure*}

\subsection{Additional Reward Details}

\paragraph{Format Reward.} Let $C$ be the number of assistant turns in a conversation, $c_i$ denote the $i^{th}$ turn response. Let $\operatorname{match}(c_i, P)\text{=}1$ if $c_i$ matches JSON tag \textless P\textgreater ...\textless /P\textgreater exactly (0 otherwise). We compute the multi-turn format reward by: 
$R_{\mathrm{fmt}}(C)\text{=}\frac{1}{C}\sum_{i=1}^{C} cot(c_i)$ (CoT), where $cot(c_i)\text{=}\operatorname{match}(c_i,\mathtt{think}) \times \bigl( \operatorname{match}(c_i,\mathtt{answer}) \oplus \operatorname{match}(c_i,\mathtt{tool\_call}) \bigr)$. $\oplus$ denotes exclusive-or. 

For DA, the formulation is: $R_{\mathrm{fmt}}(C)\text{=}\frac{1}{C}\sum_{i=1}^{C} da(c_i)$ (DA), where $da(c_i)\text{=}\operatorname{match}(c_i,\mathtt{answer}) \oplus \operatorname{match}(c_i,\mathtt{tool\_call})$, without the need of $<\text{think}>$ tag.

\paragraph{Clarification of augmented model selection on $R_{mat}$.}
We set most rows in $M_Q$ to $M_{o_i}$ because TAR and FNIR depend on operating thresholds that are selected from \emph{dataset-level} scores, rather than being determined by a single query. If we set $\gamma = 1.0$ (set all rows of $M_Q$ to $M_{o_i}$), then each model combination yields exactly one metric-based reward value, which significantly reduces exploration. In contrast, the Rank-1 outcome for a given $q$ can be assessed query-wise, whereas TAR/FNIR are much more sensitive to the global threshold selection. Therefore, we introduce $\gamma$ to control this augmentation and balance stable threshold estimation with sufficient exploration.

\subsection{Tool Design.}

Tool design should be in an efficient way for agent learning. If the number of tools or the number of parameters of tools is large, it will become more difficult for the agent to execute the tools. In our scenario, the goal of the agent is to call different biometric recognition models to extract features based on the input. Therefore, we design a universal tool that is suitable for every biometric model.  The tool takes sequential images, the biometric model name as the input, and returns the similarity matrix and the predicted label of the images. 

\paragraph{Tool Results.} \ours only receives the predicted identity label from the tool execution, while the score vectors and selected models are stored within the system. We do not return the similarity scores or confidence weights of the predicted identity, as doing so was found to prematurely halt the exploration of model combinations during training. This design choice is justified because a high confidence or similarity score from one model does not preclude further performance gains through fusion with other models, given that each model contributes independently.

\paragraph{Error Handling.}
When the tool calling fails, for example, calling a wrong model name or an invalid JSON format. It is important to let the agent know the reason for the failure during training. We design error handling for unexpected behaviors and enable the agent to identify the reason for the failures.

\section{Additional Implementation Details}

\begin{table}[t!]
\tabcolsep=0.1cm
\centering
\resizebox{\linewidth}{!}{
\begin{tabular}{lcccc}
\toprule
Dataset & Type  & \#Subjects (Train/Test/Non-mated) & \#Query & \#Gallery \\
\midrule
CCVID  & Video & 75 / 151 / 31 & 834  & 1074  \\
MEVID  & Video & 104 / 54 / 11 & 316 & 1438  \\
LTCC   & Image & 77 / 75 / 15 & 493 & 7050  \\
\bottomrule
\end{tabular}
}
\caption{Statistics of the evaluation set of human recognition benchmarks. The number of query and gallery indicate the number of images/sequences for image/video datasets.}
\label{tab:stat_datasets}
\end{table}

\paragraph{Datasets.} The dataset statistics are summarized in Tab.~\ref{tab:stat_datasets}. They comprise multi-view captures and cross-modal biometric data, enabling rigorous evaluation of generalization across diverse resolutions, viewpoints, and temporal dynamics. This comprehensive benchmarking setup ensures robustness against real-world challenges such as occlusion, motion blur, and sensor heterogeneity, thereby validating the practicality of the proposed approach in unconstrained environments. During training, we only use $2,000$ samples (or medias) maximum for each dataset to perform training set score matrices.

\paragraph{Model Pools.} We follow ~\cite{zhu2025quality} to construct the same model pools: AdaFace~\cite{kim2022adaface} (ViT-Base, WebFace4M), CAL~\cite{gu2022clothes} (ResNet50, CCVID/MEVID/LTCC), BigGait~\cite{ye2024biggait} (DINOv2-Small, CCPG), AIM~\cite{yang2023good} (ResNet50, LTCC), and AGRL~\cite{wu2020adaptive} (ResNet50, MEVID), where the former denotes the model architecture and the latter indicates the training dataset. We use \textit{\{model\}\_\{dataset\}} to denote the difference of checkpoints (\ie, CAL\_CCVID and CAL\_LTCC) during training and inference. 

\paragraph{Center Features of Training Set.} 
We follow QME~\cite{zhu2025quality} to extract center features as the gallery features for the training set. We compute the center features based on the subject ID, camera ID, and clothing ID. Therefore, each subject may have multiple center features.

\paragraph{Similarity Distances of Each Model.}
We follow QME~\cite{zhu2025quality} to measure the distances between features. AdaFace~\cite{kim2022adaface}, CAL~\cite{gu2022clothes}, AIM~\cite{yang2023good}, AGRL~\cite{wu2020adaptive}, and CLIP3DReID~\cite{liu2024distilling} use cosine-similarity to measure the distance, while BigGait~\cite{ye2024biggait} uses Euclidean distance. We use Eq.~\ref{eq:euc_scores} to transform Euclidean distance into similarity scores.

\begin{table}[t!]
\tabcolsep=0.1cm
\centering
\resizebox{\linewidth}{!}{
\begin{tabular}{lccc}
\toprule
Dataset & \#Media (Full) & \#Media (10-shot) & Percentage (\%) \\
\midrule
LTCC   & 9576 & 768 & 8.0 \\
\bottomrule
\end{tabular}
}
\caption{Comparison of the full dataset size and the few-shot (10-shot) subset used for training. The percentage indicates the proportion of data used in the few-shot setting relative to the full dataset.}
\label{tab:fewshot_datasets}
\end{table}

\paragraph{Cross-domain Training.} We conduct cross-domain evaluation through zero-shot testing and few-shot training. For the few-shot setting, we adopt a 10-shot protocol (\ie, each training subject provides 10 images/videos). The same procedure is applied to extract center features and score matrices from the training set, and only the 10-shot dataset is used for \ours training. Few-shot data size for each dataset is shown in Tab.~\ref{tab:fewshot_datasets}

\paragraph{Additional In-Domain Evaluation.}

\paragraph{Accuracy From Agent Answer.} 
As shown in Table~\ref{tab:agent_acc}, the \textit{Agent} predictions are derived by selecting identity labels from the outputs of different tools, rather than directly relying on the score fusion results. Despite this discrete decision process, the agent achieves performance comparable to the score-based fusion method (\fusion). We attribute this performance improvement primarily to the use of the answer accuracy reward.  However, \fusion consistently yields higher accuracy across all datasets, demonstrating that score-level fusion effectively integrates complementary information from multiple models and provides more reliable identity estimation than selection based solely on predicted labels.
However, since the agent’s answer can only reflect label prediction accuracy—and not other metrics such as ranking quality or calibration—different evaluation criteria may be adopted depending on application requirements. This result also provides a future direction on whether MLLMs or agents can even have a better performance \wrt metric results like verification (TAR) and open-set search (FNIR).

However, we do not observe the performance gain on CCVID. We hypothesize this is due to the performance gap between the training set and the test set. CAL is trained on CCVID, but AdaFace is not. CAL has a better Rank1 result on the training set, which makes the agent more reliant on the decision on CAL.

\paragraph{Time-consuming.} Since the agent only decides the model combination, the task itself is relatively simple. Considering the demand for faster responses in practical applications, we adopt a lightweight 3B model for efficiency. As shown in Tab.~\ref{tab:fusionagent_time}, the CoT inference of \ours, including tool executions and score-fusion, takes 2.81s per sample on a H100 device, while \da takes 1.03s. For comparison, QME~\cite{zhu2025quality} takes 0.67s per sample on average, including tool executions, quality estimating, and score-fusion.

\begin{table}[t!]
    \centering
    \scalebox{0.9}{
    \begin{tabular}{c|c|c|c}
        \toprule
        Method & CCVID & MEVID & LTCC  \\
        \midrule
        \fusion & 93.4 & 79.4 & 98.9 \\
        Agent & 81.8 & 76.9 & 96.8 \\
        \bottomrule
    \end{tabular}}
    \caption{Answer accuracy (Rank 1) performance on CCVID, MEVID, and LTCC. \fusion is the Rank 1 result evaluated from the score matrix. Agent is the accuracy evaluated from the agent responses.}
    \label{tab:agent_acc}
\end{table}

\begin{table}[t!]
\centering
\resizebox{\linewidth}{!}{
\begin{tabular}{l|cccc}
\toprule
\multirow{2}{*}{\textbf{Method}} & \multicolumn{4}{c}{\textit{Time/sample (s)}} \\
& \textbf{CCVID} & \textbf{MEVID} & \textbf{LTCC} & \textbf{Avg}\\
\midrule
QME~\cite{zhu2025quality} & 0.72 & 0.66 & 0.64 & 0.67 \\
\ours (DA) & 1.20 & 0.98 & 0.91 & 1.03\\
\ours (CoT) & 2.80 & 2.43 & 3.19 & 2.81 \\
\bottomrule
\end{tabular}
}%
\caption{Time-consuming Comparison of \ours on Different Datasets. [Keys: DA=Direct answering.]}
\label{tab:fusionagent_time}
\end{table}

\paragraph{Tool-call Efficiency.}
On CCVID, MEVID, and LTCC, \ours reduces tool uses by 32.1\%, 32.1\%, and 31.3\% compared to run-all baselines, respectively.

\paragraph{Advantage of ACT.} Tab.~\ref{tab:ltcc_cross_performance} and ~\ref{tab:ablation_score_fusion_method} reveal two regimes. With a small domain gap, ACT pairs well with \ours as the anchor prior is reliable. When the prior is miscalibrated, FarSight without anchor selection is more robust. Overall, ACT can reach higher performance, whereas FarSight is more robust under uncertainty.

\subsection{Additional Ablation Experiments}

\paragraph{Confidence Weights.} 
Tab.~\ref{tab:ablation_confidence_weight} ablates the effect of confidence weights in \fusion. Compared to Z-score, Min–max normalization underperforms by $0.9\%$ points in mAP and $0.4\%$ in Rank-1, with an even larger drop observed in FNIR ($9.8\%$). Omitting confidence weighting leads to the largest FNIR degradation (62.4), confirming that confidence-aware scaling is essential in open-set search. Among the strategies, Z-score consistently yields the best overall results, reducing FNIR by more than 10\% over other variants. This advantage likely stems from its robustness to outliers, which allows for more stable calibration across heterogeneous models. These findings indicate that while confidence weighting has a limited effect on closed-set metrics (Rank-1, TAR), it plays a critical role in improving reliability under stricter false-positive constraints.

\begin{table}[t!]
\centering
\tabcolsep=0.1cm
\resizebox{0.9\linewidth}{!}{
    \begin{tabular}{l|cccc}
    \toprule
        Norm. &  Rank1 & mAP & TAR@1\%FAR & FNIR@1\%FPIR   \\
        \midrule
        \textit{None} & $75.0$ & $40.8$ &  $36.5$ & $62.4 \pm 9.2$ \\
        \textit{Min-max} & $75.1$ & $40.5$ &  $36.6$ & $60.8 \pm 10.7$ \\
        \textit{Z-score} & $75.5$ & $41.4$ & $36.5$ & $51.0 \pm 9.4$ \\ 
    \bottomrule
    \end{tabular}
    }%
    \caption{\textbf{Ablation on confidence weighting strategies in \fusion on LTCC.} Both Min-max and Z-score normalization improve over no weighting, with Z-score achieving the best overall performance and substantially reducing FNIR. [Keys: Norm.= the method for confidence weights.]}
    \label{tab:ablation_confidence_weight}
\vspace{-0.8em}
\end{table}

\paragraph{Effects of Top-k Selection.} 
As shown in Tab.~\ref{tab:ablation_topk_ltcc}, the effect of Top-k selection is consistent across all three datasets. The overall score is computed as the sum of Rank-1, mAP, and TAR, minus FNIR.
On CCVID, applying Top-k selection brings clear improvements in Rank-1 accuracy (92.3$\rightarrow$93.4), TAR (83.3$\rightarrow$85.8), and the overall score (64.6$\rightarrow$65.5), while maintaining comparable FNIR. For MEVID, the performance remains relatively stable, with a slight increase in Rank-1 (54.1$\rightarrow$54.7) but minor fluctuations in other metrics. 
Similarly, on LTCC, Top-k selection provides marginal gains in Rank-1 (75.1$\rightarrow$75.5) and overall score (25.5$\rightarrow$25.6), with negligible changes in TAR and FNIR. Overall, Top-k selection consistently achieves a more favorable trade-off and leads to a better aggregated performance across datasets, confirming its robustness and general effectiveness.

\begin{table}[t!]
\centering
\tabcolsep=0.1cm
\resizebox{0.95\linewidth}{!}{
    \begin{tabular}{c|ccccc}
    \toprule
        Top-k &  Rank1 & mAP & TAR & FNIR & Overall   \\
        \midrule
        \rowcolor{gray!30} \multicolumn{6}{c}{\textit{CCVID}} \\
        \xmark & $92.3$ & $92.5$ & $83.3$ & $9.7 \pm 1.3$ & 64.6\\
        \cmark & $93.4$ & $92.7$ & $85.8$ & $9.9\pm1.5$ & 65.5\\
        \rowcolor{gray!30} \multicolumn{6}{c}{\textit{MEVID}} \\
        \xmark & $54.1$ & $29.1$ & $35.6$ & $58.2 \pm 8.5$ & 15.2 \\
        \cmark & $54.7$ & $28.7$ & $34.9$ & $58.6\pm7.4$ & 14.9\\
        \rowcolor{gray!30} \multicolumn{6}{c}{\textit{LTCC}} \\
        \xmark & $75.1$ & $41.8$ &  $37.2$ & $51.9 \pm 9.4$ & 25.5\\
        \cmark & $75.5$ & $41.4$ & $36.5$ & $51.0\pm9.4$ & 25.6 \\
    \bottomrule
    \end{tabular}
    }%
    \caption{\textbf{Effects of Top-k selection.} Top-k selection gains a better overall performance on three datasets.}
    \label{tab:ablation_topk_ltcc}
\end{table}

\begin{figure}
    \centering
    \includegraphics[width=\linewidth]{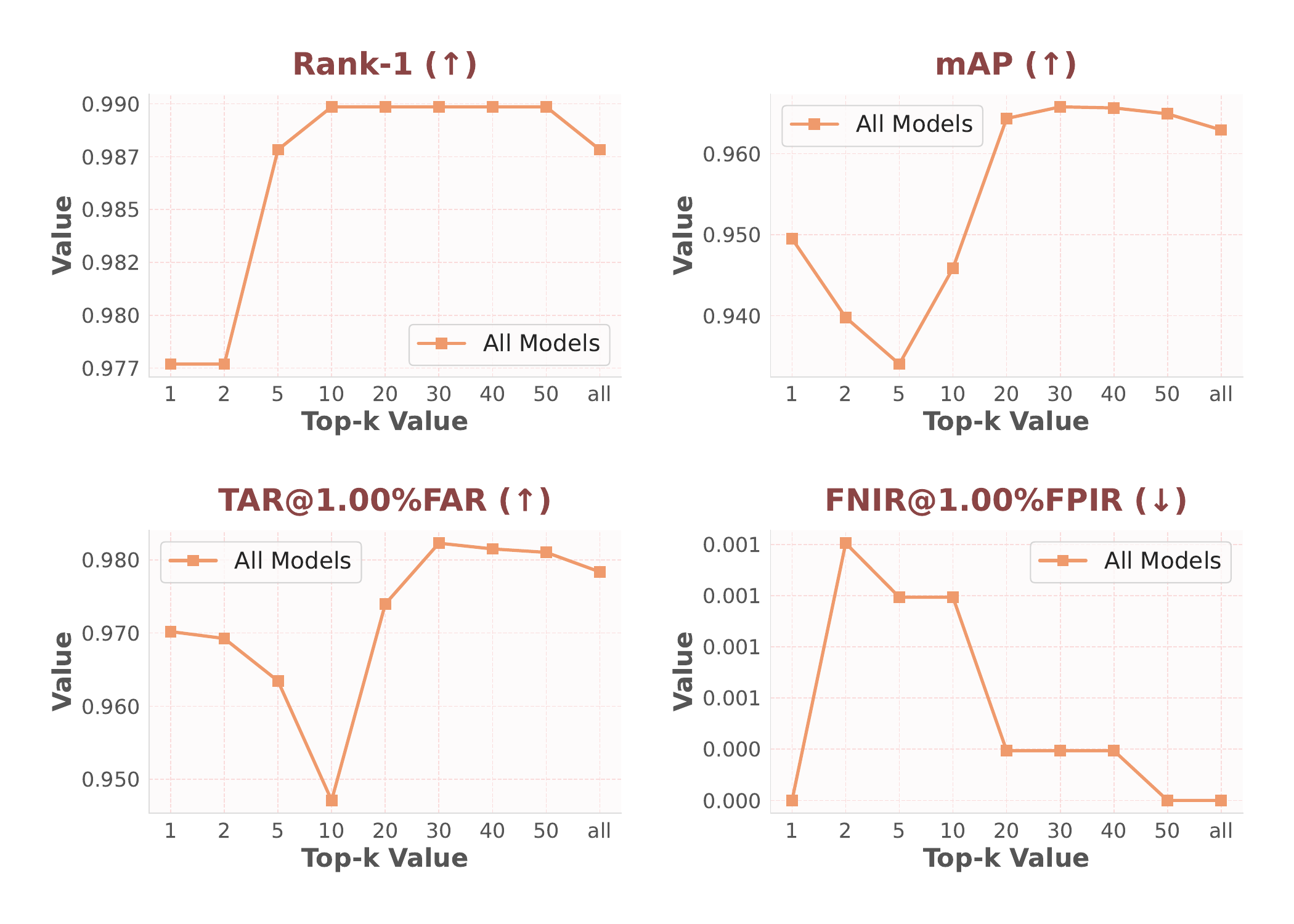}
    \caption{\textbf{Effects of Top-k Values on LTCC Training Set.} The overall performance reaches the peak when $k=40$.}
    \label{fig:ablation_topk_ltcc_trainset}
\end{figure}

\paragraph{Effects of Top-k Values on Training Set.} 
We visualize the performance comparison on the LTCC training set in Fig.~\ref{fig:ablation_topk_ltcc_trainset}. The results show that the overall performance gradually improves as $k$ increases and reaches its peak at $k=40$, after which further growth of $k$ does not yield additional benefits. Therefore, we adopt $k=40$ for the testing stage to achieve a good balance between effectiveness and stability. We follow the same strategy when selecting Top-k values for the other datasets as well.

\paragraph{Anchor Sensitivity in \fusion.} Performance of \fusion is more sensitive to anchor selection when single-model performance varies widely (Tab.~\ref{tab:rebuttal_anchor}).

\begin{table}[t]
    \centering
        \begin{tabular}{c|c}
            Anchor & Overall (CCVID) \\
            \midrule
            AdaFace & 260 \\
            BigGait &  259 \\
            CAL &   257 \\
            \midrule
              & Overall (LTCC) \\
            \midrule
            AdaFace & 77 \\
            AIM &  95 \\
            CAL &   102 \\
            \midrule
              & Overall (MEVID) \\
            \midrule
            AdaFace & 59 \\
            AGRL &  65 \\
            CAL &  65 \\
        \end{tabular}
        \caption{Anchor sensitivity.}
        \label{tab:rebuttal_anchor}
\end{table}

\subsection{Additional Qualitative Results}

\begin{figure}[t!]
    \centering
    \includegraphics[width=\linewidth]{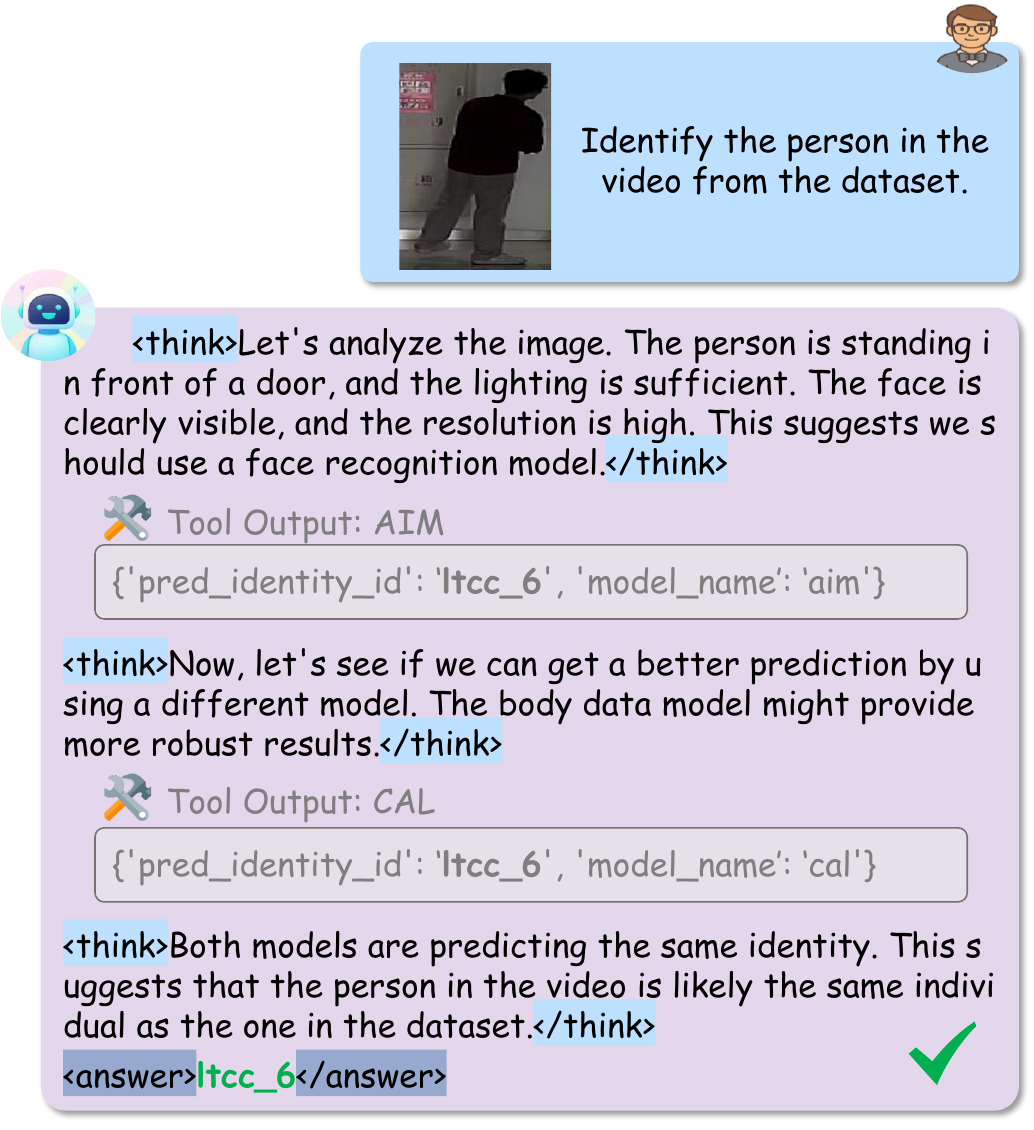}
    \caption{\textbf{CoT example of \ours on LTCC.}}
    \label{fig:supp_visualization_conv1_cot}
\vspace{-0.5em}
\end{figure}

\begin{figure}[t!]
    \centering
    \includegraphics[width=\linewidth]{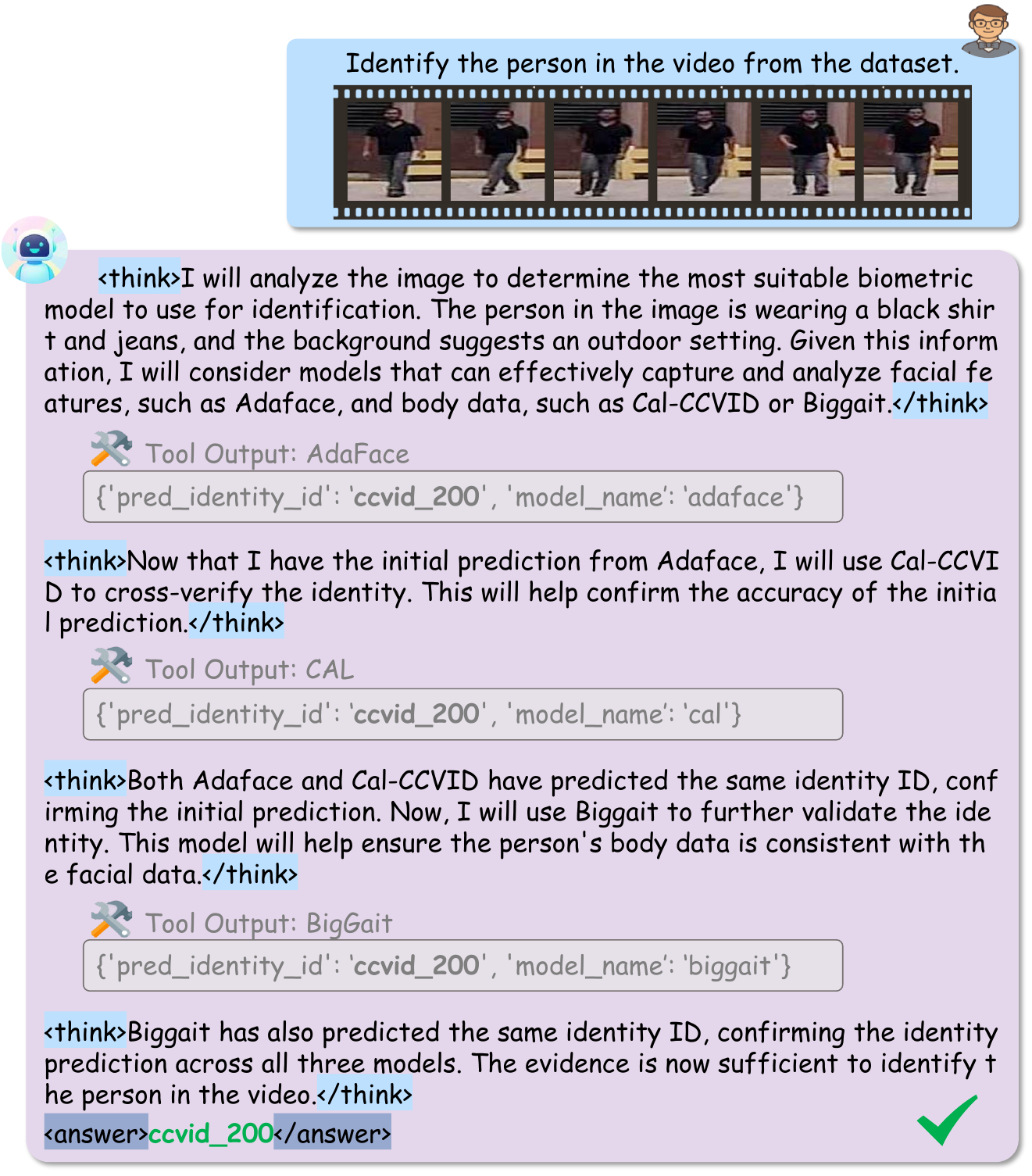}
    \caption{\textbf{CoT example of \ours on CCVID.}}
    \label{fig:supp_visualization_conv2_cot}
\vspace{-1em}
\end{figure}

\begin{figure}[t!]
    \centering
    \includegraphics[width=\linewidth]{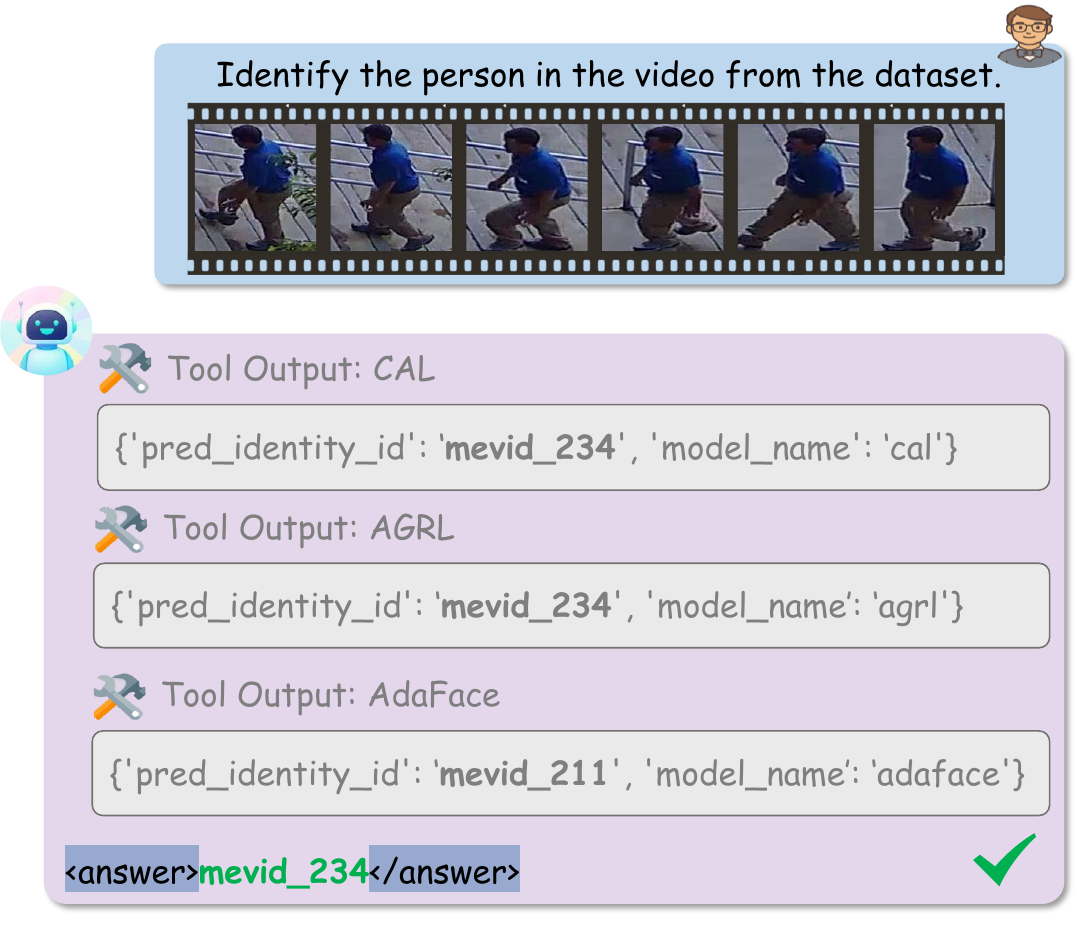}    
    \caption{\textbf{DA example of \ours on LTCC. \ours directly outputs the tool-use code and answer without thinking.}}
    \label{fig:supp_visualization_conv1_da}
\vspace{-1em}
\end{figure}

\paragraph{Additional Conversations.} 
Fig.~\ref{fig:supp_visualization_conv1_cot} and~\ref{fig:supp_visualization_conv2_cot} present additional CoT examples from \ours. Depending on the input query pattern, \ours dynamically selects the most suitable models. Fig.~\ref{fig:supp_visualization_conv1_da} illustrates the DA (direct answering) mode of \ours, which bypasses reasoning generation to accelerate inference while maintaining comparable performance.


\section{Limitations}

\paragraph{Reasoning Collapse.}
RFT may lead to unstable or degenerate reasoning behaviors during training. For instance, the agent’s reasoning content can become repetitive, disregarding the actual differences in model-predicted identities as training progresses. This phenomenon likely arises from \textit{reward hacking}, since no explicit supervision is provided on the quality of reasoning. Stabilizing the reasoning process and ensuring consistent answer quality remains an important direction for future research.

\paragraph{Model Combination Estimation.} 
In our setting, each sample can be associated with multiple possible model combinations. As the number of samples increases, the search space grows exponentially, making grid search for the ground-truth optimal combination computationally infeasible. Consequently, we adopt the proposed metric-based reward to estimate the overall performance without exhaustively enumerating all combinations. In the future, exploring more efficient or learning-based strategies for model combination estimation could further enhance scalability and accuracy.

\section{Potential Societal Impacts}

Our paper leverages multiple public biometric datasets for research purposes, with a focus on the similarity score domain, which is less directly tied to sensitive biometric data. As biometric recognition tasks grow increasingly complex, integrating multiple models has become a key trend to enhance system performance. It is essential to ensure that the use of biometric datasets and recognition systems adheres to ethical standards and complies with privacy regulations.

\end{document}